\documentclass[11pt,a4paper]{article}
\ifdefined\pdfminorversion\pdfminorversion=7\fi

\usepackage{memoraxpaper}
\usepackage[
  backend=biber,
  style=authoryear-comp,
  natbib=true,
  maxcitenames=2,
  maxbibnames=12,
  uniquename=init,
  giveninits=true,
  doi=true,
  url=true,
  isbn=false,
  eprint=true
]{biblatex}
\AtBeginBibliography{\raggedright\sloppy}

\usepackage{amsmath,amsfonts,bm}

\def\eqref#1{equation~\ref{#1}}
\def\1{\bm{1}}

\DeclareMathAlphabet{\mathsfit}{\encodingdefault}{\sfdefault}{m}{sl}
\SetMathAlphabet{\mathsfit}{bold}{\encodingdefault}{\sfdefault}{bx}{n}

\usepackage{multirow}
\usepackage{longtable}
\usepackage{wrapfig}
\usepackage[scaled=0.85]{beramono}
\usepackage[T1]{fontenc}
\usepackage{textcomp}
\usepackage[most]{tcolorbox}
\usepackage{xcolor}
\usepackage{listings}
\usepackage{enumitem}
\usepackage{needspace}
\makeatletter

\definecolor{pbFrame}{HTML}{2B3A4A}
\definecolor{pbBack}{HTML}{F7F8FA}
\definecolor{pbAccent}{HTML}{1F4E79}
\definecolor{pbMuted}{HTML}{6B7280}
\definecolor{pbRule}{HTML}{C5CBD3}

\newcommand{\promptfont}{\small}
\newcommand{\jsonfont}{\footnotesize\ttfamily}

\newtcolorbox[auto counter, number within=section]{promptbox}[2][]{%
  breakable, enhanced, sharp corners,
  colback=pbBack, colframe=pbFrame, boxrule=0.7pt,
  left=8pt, right=8pt, top=6pt, bottom=6pt,
  fonttitle=\bfseries\small, coltitle=white, colbacktitle=pbFrame,
  fontupper=\promptfont,
  before upper={\setlength{\parskip}{3pt}},
  lines before break=4,
  title after break={Box~\thetcbcounter\ (continued)},
  before skip=9pt, after skip=9pt,
  pad at break*=3pt,
  title={Box~\thetcbcounter: #2}, #1
}

\newcommand{\pb@sechead}[1]{%
  \noindent{\bfseries #1}\par\nobreak
  \vskip 3pt
  {\color{pbRule}\hrule height 0.4pt}%
  \nobreak\vskip 4pt
  \noindent}
\newcommand{\promptsec}[1]{\par\addvspace{7pt}\pb@sechead{#1}}
\newcommand{\promptsectop}[1]{\par\pb@sechead{#1}}

\newtcolorbox{promptaskbox}{%
  blanker, breakable,
  left=9pt, right=6pt, top=3pt, bottom=3pt,
  before skip=4pt, after skip=4pt,
  borderline west={1.6pt}{0pt}{pbAccent}}
\newcommand{\promptask}[1]{\begin{promptaskbox}\itshape #1\end{promptaskbox}}

\newcommand{\ptype}[1]{\textbf{#1}}

\newenvironment{pitemize}
  {\begin{itemize}[leftmargin=1.15em,labelsep=0.4em,itemsep=2pt,
                   topsep=2pt,parsep=0pt,partopsep=0pt,beginpenalty=10000]}
  {\end{itemize}}
\newenvironment{penumerate}
  {\begin{enumerate}[leftmargin=1.35em,labelsep=0.4em,itemsep=2pt,
                     topsep=2pt,parsep=0pt,partopsep=0pt,beginpenalty=10000]}
  {\end{enumerate}}

\makeatother

\lstdefinestyle{promptjson}{%
  basicstyle=\jsonfont,%
  columns=fullflexible,%
  keepspaces=true,%
  upquote=true,%
  breaklines=true,%
  breakatwhitespace=true,%
  breakindent=1.2em,%
  showstringspaces=false,%
  frame=none,%
  xleftmargin=0pt,%
  aboveskip=0pt,%
  belowskip=0pt,%
  alsoletter={_},%
  morekeywords={base_task,historical_pr,repo,problem_statement,changed_files,patch,title,body,state,is_merged,labels,diff_hunks,commit_messages,label,relation_type,relationship,confidence,brief_reason,evidence_spans,boundary_note},%
  keywordstyle=\color{pbAccent},%
  moredelim=[s][\color{pbMuted}]{<}{>}%
}

\newtcblisting{jsonbox}[1]{%
  unbreakable, enhanced, sharp corners,
  colback=pbBack, colframe=pbFrame, boxrule=0.7pt,
  left=7pt, right=6pt, top=4pt, bottom=4pt, boxsep=2pt,
  fonttitle=\bfseries\small, coltitle=white, colbacktitle=pbFrame,
  before skip=9pt, after skip=9pt,
  pad at break*=3pt,
  listing only, listing style=promptjson,
  title={#1}
}

\newcommand{\authorentry}[2]{#1\textsuperscript{#2}}

\reporttitle{MemDev-Bench: Does Memory Help Code Agents?}
\reportsubtitle{Benchmarking Memory Systems over Real Repository Histories}
\reportrunningtitle{MemDev-Bench}
\reportauthors{\texorpdfstring{% Three centered rows in the supplied author order; no contribution roles.
\begingroup
\centering
\fontsize{11}{13.2}\selectfont
\setlength{\tabcolsep}{0pt}
\setlength{\parskip}{0pt}
\renewcommand{\arraystretch}{1.0}
\begin{tabular}{@{}c@{\hspace{5mm}}c@{\hspace{5mm}}c@{\hspace{5mm}}c@{}}
\authorentry{Guorui Chen}{1} &
\authorentry{Tao Liu}{1} &
\authorentry{Zhenzhen Huang}{1} &
\authorentry{Yidan Zhang}{1} \\
\authorentry{Liyu Jiang}{1} &
\authorentry{Shuoyuan Wang}{2} &
\authorentry{Hanchen Sun}{1} &
\authorentry{Xu He}{3} \\
\authorentry{Dong Li}{3} &
\authorentry{Yihan Hu}{3} &
\authorentry{Jianye Hao}{3} &
\authorentry{Lei Feng}{1}
\end{tabular}\par
\endgroup
}{Guorui Chen; Tao Liu; Zhenzhen Huang; Yidan Zhang; Liyu Jiang; Shuoyuan Wang; Hanchen Sun; Xu He; Dong Li; Yihan Hu; Jianye Hao; Lei Feng}}
\reportaffiliations{%
  \textsuperscript{1}Southeast University, Nanjing, China\\
  \textsuperscript{2}Southern University of Science and Technology, Shenzhen, China\\
  \textsuperscript{3}MemoraX AI, Shenzhen, China%
}
\reportversion{}
\reportrunningversion{}
\reportdate{}
\reportcontact{}
\reportfooter{MemoraX AI Research · MemDev-Bench}
\reportlogo{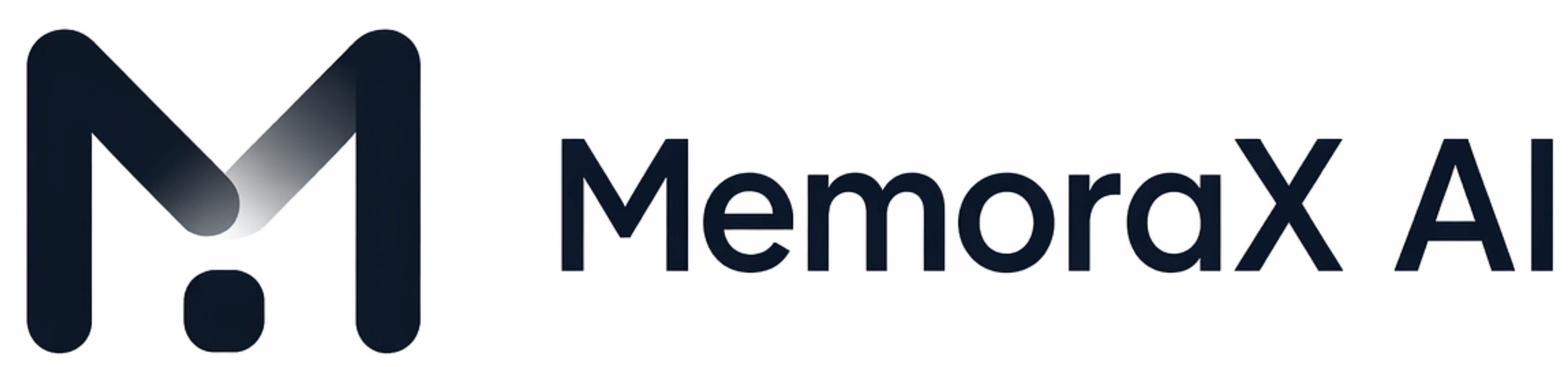}
\reportsubject{Evaluating memory systems for code agents over real repository histories}
\reportkeywords{code agents, memory systems, software engineering, repository histories}

\AtBeginDocument{%
  \hypersetup{%
    pdftitle={MemDev-Bench: Does Memory Help Code Agents? Benchmarking Memory Systems over Real Repository Histories},
    pdfauthor={Guorui Chen; Tao Liu; Zhenzhen Huang; Yidan Zhang; Liyu Jiang; Shuoyuan Wang; Hanchen Sun; Xu He; Dong Li; Yihan Hu; Jianye Hao; Lei Feng}%
  }%
}

\begin{document}

\makememoraxpapertitle

\begin{reportabstract}
Code agents increasingly work on evolving real repositories, and memory systems aim to help them reuse earlier experience.
However, such experience is diverse and evolving, so it remains unclear how memory systems organize it and thereby affect code agents.
Existing benchmarks cannot answer this reliably because they lack either temporally valid histories or control over prior-task relevance.
In this paper, we present MemDev-Bench to evaluate how memory systems affect code agents over diverse repository histories. It pairs each task with relevance-annotated earlier tasks from the same repository to build histories with controlled relevance. 
We also build an automated pipeline that reliably makes these earlier tasks executable and verifiable, producing 4,869 historical tasks for 555 base tasks.
Based on MemDev-Bench, our systematic evaluation of memory systems across diverse development scenarios yields three main findings.
(1) Memory mainly improves fine-grained localization and editing, increasing functional correctness. (2) Memory also causes failures on otherwise solvable tasks even with related history, and unrelated history tends to increase such failures. These failures mainly stem from agents reusing memory without reassessing its applicability. (3) Compared with using memory systems, directly providing agents with raw historical trajectories substantially reduces their steps. 
Further analysis suggests that agents save these steps by reusing process information that compressed memories tend to omit.
These results suggest that future memory systems should preserve process information and reassess memory applicability against current task context.
\end{reportabstract}

\printreportkeywords

\section{Introduction}
\label{sec:introduction}
Code agents powered by large language models (LLMs) have achieved substantial progress on real-world software engineering tasks, such as bug fixing and feature development.
Despite this progress, existing agents typically solve each task in isolation and start from scratch.
In contrast, human developers continuously accumulate experience from past issues and reuse it to resolve new problems efficiently.
Memory systems~\citep{Mem0,MemOS,AMem,SuperMemory,ClaudeMem} offer a way to close this gap, allowing code agents to retain repository context and problem-solving experience across tasks.
However, real repository histories mix valuable experience with abundant irrelevant information.
This raises a critical question: do memory systems consistently benefit code agents across long-term repository development?

Addressing this question requires a benchmark with two properties: temporally valid histories from real repositories and controlled relevance between prior and target tasks.
Temporal validity requires each prior task to precede the target task without revealing its solution; otherwise, memory gains may come from answer leakage rather than experience reuse.
Controlled relevance specifies whether a prior task can help the target task, so a benchmark can separate ineffective memory use from missing useful experience and test whether irrelevant history misleads agents. To our knowledge, no existing benchmark satisfies both properties.
Session-based benchmarks build histories from synthetic multi-session scenarios~\citep{dreambench_swe} or from task streams of long-term workflows~\citep{contextweave}, rather than from real repository evolution.
Repository-based benchmarks instead build tasks from GitHub histories.
Strict temporal filtering is not always applied, so histories may include tasks that postdate the target task. Such histories may therefore leak target solutions~\citep{swecontextbench}.
Other benchmarks preserve temporal order.
But they either overlook actual task relevance~\citep{swebench_cl} or use chains of almost entirely interdependent tasks~\citep{chainswe}.
Both designs make it difficult to assess how memory systems behave across diverse development histories.

We therefore propose MemDev-Bench, a benchmark that evaluates how memory systems affect code agents' task-solving behavior across diverse repository histories.
Each instance pairs a base task with relevance-annotated historical tasks from the same repository that precede it and do not reveal its solution.
We take base tasks from SWE-bench Verified and SWE-bench Pro~\citep{swebench_pro}, mine earlier pull requests (PRs) through references and code history, and label each as Strong Related, Weak Related, or Unrelated based on how much it helps solve the base task.
Following SWE-bench~\citep{swebench}, we convert these PRs into executable, verifiable tasks. 
Each PR's patch is split into a solution patch that implements the fix and a test patch that verifies it.
The environment is rebuilt from the repository state and dependencies at that time.
To make this conversion more reliable, we build a new automated pipeline.
Existing pipelines can misclassify test and solution changes or miss patch-induced environment changes, making task validation unreliable~\citep{swebench_live,repo2rlenv,swefactory}.
To address these issues, our pipeline splits patches using repository test configurations and AST analysis. It also prepares the pre- and post-fix environments separately to capture patch-induced changes.
Some pipelines rely heavily on agents for environment planning and debugging, making construction costly at scale~\citep{menvagent,sweuniverse}.
We instead use a rule--experience--agent hierarchy that invokes an agent only when rules and validated experience fail.
With this pipeline, we build 4,869 executable historical tasks for 555 base tasks across 22 repositories and 4 languages.

Using MemDev-Bench, we build three history scenarios: Relevant History with only related tasks, Noisy History with added unrelated tasks, and Repo History with longer-term development. 
In each scenario, the agent solves historical tasks in chronological order, and the memory system builds memory from trajectories. 
The agent then solves the base task with this memory. 
Based on these scenarios, we systematically evaluate representative memory systems, yielding three main findings: (i) Across all three scenarios, memory consistently helps solve tasks that the agent fails without it. These gains come mainly from finer-grained localization and editing within files. 
(ii) Memory also causes failures on tasks that the agent solves without it, even with related history, and unrelated history tends to increase them.
(iii) Compared with using memory systems, directly providing raw historical trajectories to agents substantially reduces their steps. 
Further analysis suggests that this reduction comes from process information preserved in raw trajectories but often omitted by compressed memories. 
These results suggest that future memory systems should preserve process information and reassess memory applicability against current task context.

% In summary, our contributions are threefold:
% \begin{itemize}[leftmargin=1.2em,labelsep=0.45em,itemsep=0pt,topsep=1pt,parsep=0pt,partopsep=0pt]
%     \item MemDev-Bench, which builds relevance-controlled history scenarios of different complexity.
%     \item An automated pipeline that accurately converts historical PRs into reproducible executable tasks.
%     \item The first systematic evaluation of memory systems for code agents over real repository histories.
% \end{itemize}

\section{MemDev-Bench: Construction and Statistics}
\label{sec:bench}

MemDev-Bench evaluates how memory systems affect code agents on later tasks after these systems build memory from agent trajectories on earlier tasks in the same repository.
Each evaluation case pairs a software engineering task, referred to as the base task, with its historical tasks. 
We collect base tasks from SWE-bench Verified~\citep{swebench_verified} and SWE-bench Pro~\citep{swebench_pro}, and refer to the pull request (PR) that resolves a base task as its base PR. 
For each base task, we mine earlier PRs from the same repository and annotate how much each PR helps solve the base task. 
Each annotated PR is then converted into an executable historical task in a reproducible Docker environment, which restores the repository state and dependencies at that point in history.

\begin{figure}[!t]
\centering
\includegraphics[width=0.80\linewidth]{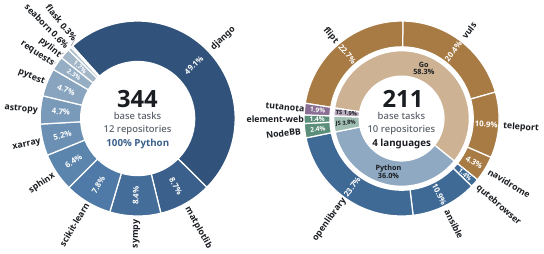}
\caption{\textbf{Base tasks distribution.}
Left: SWE-bench Verified (344 tasks, 12 Python repositories). Right: SWE-bench Pro (211 tasks, 10 repositories, four languages).}
\label{fig:base_tasks}
\end{figure}

MemDev-Bench contains 555 base tasks and 4{,}869 relevance-annotated historical tasks from 22 real-world GitHub repositories, spanning Python, Go, JavaScript, and TypeScript. Figure~\ref{fig:base_tasks} shows how these tasks are distributed over repositories and languages.
 
\subsection{Candidate Historical PR Mining}
\label{sec:mining}

\begin{figure}[!t]
\centering
\includegraphics[width=1.0\linewidth]{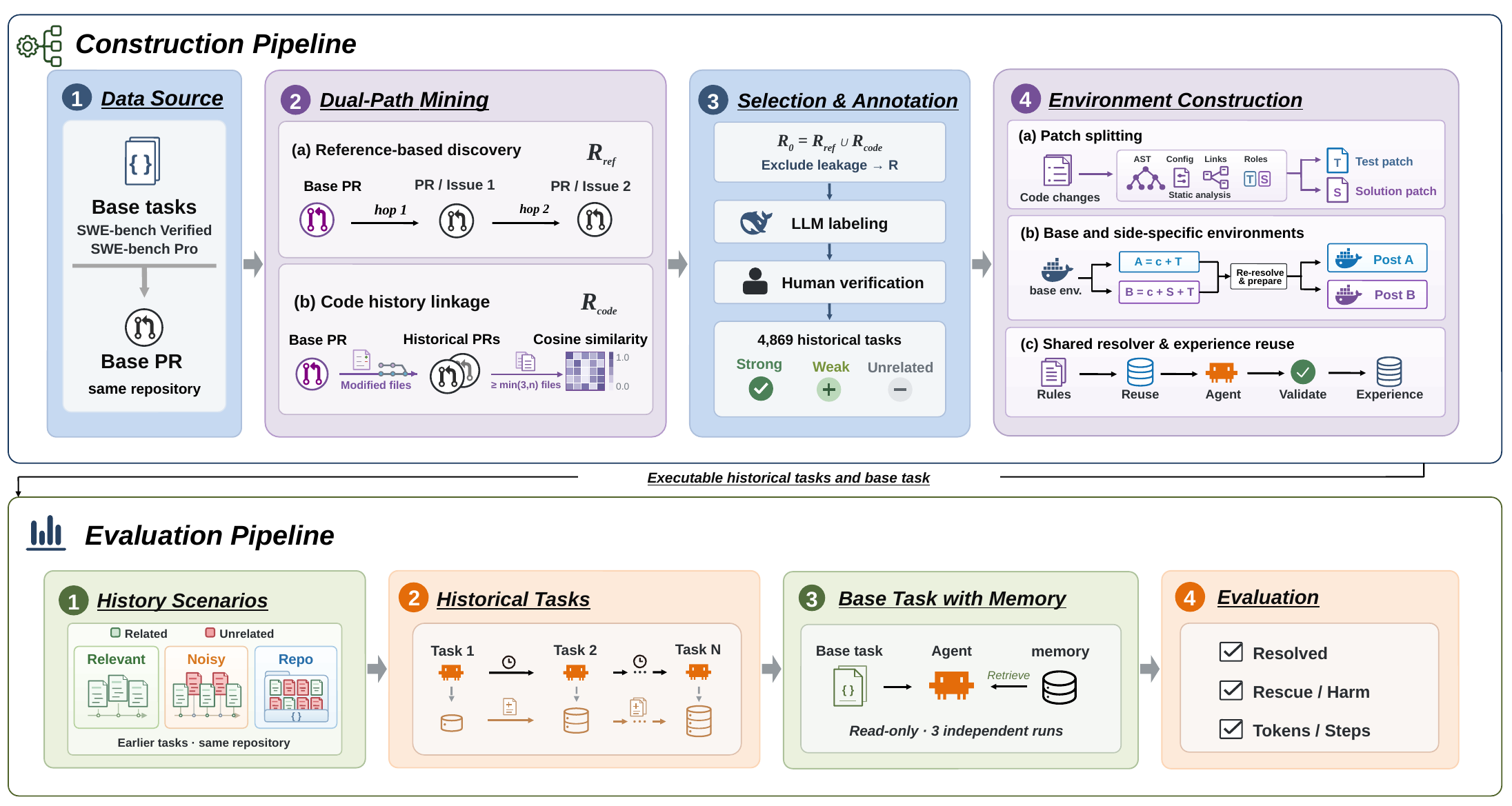}
\caption{\textbf{Benchmark construction and evaluation pipeline.}
\textit{Top:} we mine candidate PRs, annotate their relevance, and convert them into executable historical tasks.
\textit{Bottom:} an agent solves historical tasks chronologically to build memory, then solves the base task with the frozen memory.}
\label{fig:pipeline}
\end{figure}

A useful set of historical tasks should include both tasks that genuinely help with the base task and other development tasks, reflecting the interleaved development of a real repository. 
Taking every PR before the base PR would give such a set, but real repositories hold thousands of PRs, and annotating them all and building their environments would be prohibitively expensive. 
We therefore mine candidate PRs potentially relevant to the base task along two complementary routes: explicit references and code history, as shown in the Dual-Path Mining stage of Figure~\ref{fig:pipeline}. 
Temporal order is enforced using each candidate PR's terminal timestamp.
A merged candidate must be merged before the base PR is created, while an unmerged candidate must be closed before the base PR is created.

\textbf{Reference-based discovery.}
Developers often refer to earlier issues and PRs in PR descriptions and linked issues, and these referenced tasks often depend on or share designs with the current task. As shown in panel (a) of the Dual-Path Mining stage in Figure~\ref{fig:pipeline}, we extract such references from the base PR and its linked issues, then expand once more from the resulting issues and PRs, yielding PRs up to two hops away. Those from the same repository form the candidate set $R_{\mathrm{ref}}$.

\textbf{Code-history-based discovery.}
Explicit references capture only the relationships that developers choose to record. 
Many other historical PRs are related to the base PR at the code level, but they are not explicitly linked. 
Panel (b) of the Dual-Path Mining stage in Figure~\ref{fig:pipeline} shows how these PRs are found. 
Starting from each file modified by the base PR, we trace its commit history and map the earlier commits to their PRs. Frequently modified files such as configuration and utility files can create accidental overlap between otherwise unrelated PRs. 
To reduce such overlap, the mapped PRs pass through two filters. 
The first filter keeps PRs that share at least $\min(3,n)$ changed files with the base PR, where $n$ is the number of files changed by the base PR. 
The second filter encodes each PR with the MiniMax embo-01 embedding model, using its title, body, changed file paths, commit messages, and code diffs as input. 
It then keeps PRs whose cosine similarity to the base PR exceeds $0.65$, and these PRs form $R_{\mathrm{code}}$. The two routes form a preliminary pool $R_0 = R_{\mathrm{ref}} \cup R_{\mathrm{code}}$. We screen $R_0$ for solution leakage, such as backports and near-duplicate fixes. Human reviewers then exclude historical tasks that directly reveal the base solution. The remaining PRs form the candidate pool $R$ for relevance annotation.

\subsection{Candidate Historical PR Annotation}
\label{sec:anno}

\textbf{Relevance levels.}
After building the candidate pool $R$, we assign each candidate one of three relevance levels. A \textbf{Strong Related} PR provides direct and actionable help for the base task, such as a reusable implementation or the same root cause. A \textbf{Weak Related} PR provides relevant implementation or testing context, but its concrete solution cannot transfer to the base task. An \textbf{Unrelated} PR provides no substantive help, even if it shares files or code with the base PR. Strong/Weak Related PRs include merged PRs only, since accepted changes provide more reliable experience for later tasks. Unrelated PRs also include some unmerged PRs, which reflect the failed attempts and incomplete records of real development history. We further assign each candidate one of 20 fine-grained relation types that characterize its relationship to the base task. Appendix~\ref{app:fine-grained_relation_types} details these types.

\textbf{Annotation procedure.}
To annotate the candidates, we first use DeepSeek-V4-Pro~\citep{deepseek_v4} to label each candidate based on the textual descriptions and code changes of both the candidate and the base task. Appendix~\ref{app:annotation} details the LLM annotation setup. Next, three senior software engineers with extensive development experience in the benchmark languages verify these annotations. Samples are partitioned among them and exchanged so that every sample is examined by all three, and disagreements are resolved through discussion. After cross-checking, $87.82\%$ of the LLM annotations are retained,  while the remainder are replaced with the human consensus labels. Because this verification exposes annotators to the LLM annotations, another three engineers independently label $2{,}270$ samples from scratch, covering all Strong Related and Weak Related samples plus 500 randomly sampled Unrelated cases. These blind annotations reach Fleiss' $\kappa$~\citep{fleiss1971} of $0.88$ for relevance level and $0.80$ for fine-grained relation type. They match the main-verification labels for $97.1\%$, $92.0\%$, and $96.2\%$ of Strong Related, Weak Related, and Unrelated samples, respectively. Remaining disagreements are resolved through discussion to obtain the final labels.

\subsection{Executable Historical Task Construction}
\label{sec:task_construction}
After annotation, we convert each candidate PR into an executable historical task, which includes a reproducible environment and its solution and test patches.
Prior pipelines face three key issues in environment construction~\citep{swebench,swebench_live,repo2rlenv,swefactory,menvagent,sweuniverse}. As shown in the Environment Construction stage of Figure~\ref{fig:pipeline}, our pipeline addresses them with three components.
First, prior pipelines mainly split test and solution patches by file paths or names, which can misclassify changes when repositories use custom test layouts. To avoid this, our patch splitter combines repository-specific test configurations with AST-based static analysis to separate the two patches according to code dependencies.
Second, many pipelines reuse one fixed environment recipe after applying the patch and thus miss environment changes introduced by the patch. We therefore prepare the pre- and post-fix environments separately, so these changes no longer make correct fixes fail or incorrect fixes pass.
Third, pipelines that rely heavily on agents for planning and debugging are costly at scale. We use a resolver with a rule--experience--agent hierarchy, invoking an agent only when deterministic rules and validated fixes from earlier tasks both fail. Validated agent fixes are then stored for later tasks with matching conditions.
Appendix~\ref{app:environment_construction} details each component and compares these designs with prior methods.

\subsection{Historical Task Statistics}
\label{sec:stats}

Through the above process, we construct MemDev-Bench with 555 base tasks and 4,869 associated historical tasks. The SWE-bench Verified and Pro base tasks are associated with 2,987 and 1,882 historical tasks, averaging 8.68 and 8.92 per base task, respectively. We next analyze these tasks by relevance level, time gap, and change size.

\begin{figure}[t]
\centering
\includegraphics[width=\linewidth]{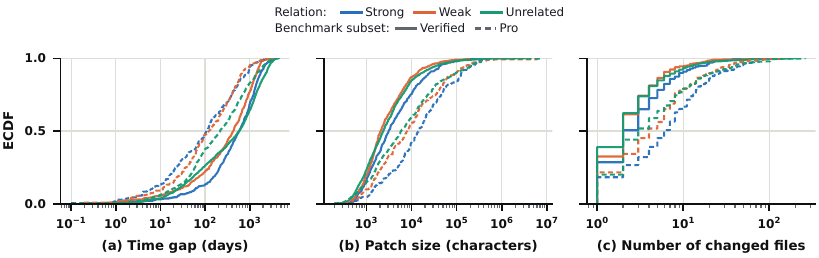}
\caption{\textbf{Empirical cumulative distributions of historical task properties.}
(a) Time gap to the base task;
(b) non-test patch size in characters;
(c) number of changed files.}
\label{fig:properties}
\end{figure}

% Keep the relevance summary beside its explanatory paragraph.
\Needspace{10\baselineskip}
\begingroup
\setlength{\intextsep}{0pt}
\setlength{\columnsep}{14pt}
\begin{wraptable}[6]{r}{0.40\textwidth}
\centering
\footnotesize
\captionsetup{font=footnotesize,skip=4pt}
\renewcommand{\arraystretch}{1.0}
\setlength{\tabcolsep}{4pt}
\caption{\textbf{Relevance composition} (\%).}
\label{tab:label_dist}
\begin{tabular}{@{}lrr@{}}
\toprule
Relevance level & Verified & Pro \\
\midrule
Strong Related & 19.6 & 13.5 \\
Weak Related & 20.5 & 16.9 \\
Unrelated, merged & 47.0 & 65.2 \\
Unrelated, unmerged & 12.9 & \phantom{0}4.4 \\
\bottomrule
\end{tabular}
\end{wraptable}

\noindent\textbf{Label composition.} As shown in Table~\ref{tab:label_dist}, Strong Related and Weak Related tasks, which provide information useful for the base task, account for $40.1\%$ and $30.4\%$ of the historical tasks in the Verified and Pro subsets, respectively. Unmerged PRs appear only in Unrelated, reflecting failed attempts and incomplete historical records.
\par\smallskip

\noindent\textbf{Time gap.} As illustrated in Figure~\ref{fig:properties}a, the gap between a historical task and its base task ranges from under a day to several years at every relevance level in both subsets. MemDev-Bench therefore covers both recent and long-term development history.
\par
\endgroup

\noindent\textbf{Change size.} As illustrated in panels (b) and (c) of Figure~\ref{fig:properties}, non-test patch size and changed-file count span a wide range at every relevance level. The tasks range from compact local edits to large changes across many files, with SWE-bench Pro tasks involving larger changes.

Together, these distributions show that historical tasks vary substantially in relevance, temporal distance, and change scale. This diversity enables MemDev-Bench to construct evaluation scenarios that better reflect the histories encountered in real repository development.

\section{Experimental Setup}
To control evaluation cost, all experiments use the Verified subset of MemDev-Bench, which contains 344 base tasks and 2{,}987 historical tasks from 12 Python repositories.

\label{sec:setup}
\textbf{Agent and base model.} We use Claude Code~\citep{claude_code} as the agent and DeepSeek-V4-Flash~\citep{deepseek_v4} as the base model.
Automatic context compaction is disabled to avoid affecting the measured effect of memory.
Configurations remain fixed throughout the evaluation; Appendix~\ref{app:agent_config} provides full details and hyperparameters.
 
\textbf{Information isolation.}
Measuring the effect of memory requires the agent to solve each task on its own. 
However, SWE-bench tasks come from public GitHub repositories, and their original PRs and fix patches remain available online. 
When web access is unrestricted, the agent often retrieves them and resolves $93.31\%\pm0.87\%$ of the 344 SWE-bench Verified tasks even without memory. 
Such answer leakage makes the resolve rate reflect answer retrieval rather than task solving, and leaves limited headroom for measuring the effect of memory. 
We therefore apply an information-isolation protocol to all experiments. 
The prompt prohibits the agent from searching for task solutions. In addition, tool-level rules block sources that may reveal answers, such as search engines and public code-hosting platforms, while keeping the tools needed for normal task solving. 
Under this protocol, the resolve rate drops to $73.16\%\pm0.61\%$, which better reflects the agent's genuine task-solving ability. 
Appendix~\ref{app:isolation} gives the full prompt, the blocking rules, and the tools that remain available.

\textbf{Task subsets and history scenarios.}
The three no-memory runs under information isolation define two task subsets based on baseline consistency.
\textbf{Improvement Tasks} comprise the 133 tasks that fail in at least two runs and evaluate whether memory can solve otherwise failed tasks.
From the 211 tasks that succeed in at least two runs, we sample 50 \textbf{Preservation Tasks} covering all SWE-bench Verified repositories to evaluate whether memory preserves task-solving ability while controlling evaluation cost.
Each subset is evaluated under three history scenarios based on the relevance annotations.
\textbf{Relevant History} contains Strong Related and Weak Related tasks (3--10), representing continued development around related features or code regions.
\textbf{Noisy History} additionally includes Unrelated tasks (7--21), representing interleaved related and unrelated work.
\textbf{Repo History} reflects longer-term development. It keeps all Strong Related and Weak Related tasks and adds the most recent earlier tasks associated with other base tasks in the same repository. 
To control cost and avoid exceeding the context window, we cap each history at 35 tasks, giving 13--35 tasks per base task.

\textbf{Baselines and memory systems.} We use two baselines. \textbf{NoMem} provides no historical information and serves as the main reference; its results come from three fresh runs on each subset, independent of the runs used for task selection. \textbf{FullTraj} places raw historical trajectories into the agent context in chronological order, without compression or selection, giving a long-context reference for assessing the benefit of memory organization. Against these baselines, we evaluate five representative systems that integrate with Claude Code and cover academic methods, open-source systems, and mature production systems: A-Mem~\citep{AMem}, Mem0~\citep{Mem0}, MemOS~\citep{MemOS}, Supermemory~\citep{SuperMemory}, and Claude-Mem~\citep{ClaudeMem}. For broader coverage, Appendix~\ref{app:all_systems} reports results for 29 memory systems under Relevant History.

\textbf{Evaluation protocol.}
For each memory system, task subset, and history scenario, we follow the
evaluation pipeline in Figure~\ref{fig:pipeline}: the agent solves historical
tasks in chronological order while the memory system generates, stores, and
organizes memory. The resulting memory store is then frozen and used read-only
for three base-task runs with distinct random seeds, preventing information
from the current task or one run from affecting another.

\providecommand{\refsys}[1]{\textit{#1}}
\providecommand{\sd}[1]{\textcolor{black!45}{\scalebox{0.80}{$\pm$#1}}}
\providecommand{\ci}[1]{\textcolor{black!45}{\scalebox{0.80}{#1}}}
\FloatBarrier
\section{Results and Analysis}
\label{sec:results}
 
\subsection{End-to-End Performance}
\label{sec:results_overall}
 
Tables~\ref{tab:improvement_results} and~\ref{tab:preservation_results} compare end-to-end performance across systems and history scenarios. Resolved measures task success; Rescue and Harm quantify gains and regressions relative to NoMem; Input Tokens, Output Tokens, and Agent Steps measure solving cost. Appendix~\ref{app:main_metrics} defines the metrics.

\begingroup
\begin{table}[!htbp]
\centering
\caption{Results on 133 \textbf{Improvement Tasks} under Relevant (Rel.),
Noisy (Noi.), and Repo history. Second lines give three-run SDs, or $95\%$
bootstrap CIs for Rescue/Harm. Bold marks the best non-NoMem result; lower is
better for Harm and cost. Appendix~\ref{app:main_metrics} defines the metrics.}
\label{tab:improvement_results}
\scriptsize
\setlength{\tabcolsep}{1.6pt}
\renewcommand{\arraystretch}{0.95}
\resizebox{\linewidth}{!}{%
\begin{tabular}{l ccc @{\hskip 4pt} ccc @{\hskip 4pt} ccc @{\hskip 4pt} ccc @{\hskip 4pt} ccc @{\hskip 4pt} ccc}
\toprule
& \multicolumn{3}{c}{Resolved (\%)}
& \multicolumn{3}{c}{Rescue (\%)}
& \multicolumn{3}{c}{Harm (\%)}
& \multicolumn{3}{c}{Tok.\ In (M)}
& \multicolumn{3}{c}{Tok.\ Out (K)}
& \multicolumn{3}{c}{Steps} \\
\cmidrule(lr){2-4}\cmidrule(lr){5-7}\cmidrule(lr){8-10}\cmidrule(lr){11-13}\cmidrule(lr){14-16}\cmidrule(lr){17-19}
\textbf{System} & Rel. & Noi. & Repo & Rel. & Noi. & Repo & Rel. & Noi. & Repo & Rel. & Noi. & Repo & Rel. & Noi. & Repo & Rel. & Noi. & Repo \\
\midrule
\refsys{NoMem} & \multicolumn{3}{c}{35.3\,\sd{2.7}} & \multicolumn{3}{c}{--} & \multicolumn{3}{c}{--} & \multicolumn{3}{c}{6.45\,\sd{0.23}} & \multicolumn{3}{c}{59.5\,\sd{3.6}} & \multicolumn{3}{c}{185.1\,\sd{5.3}} \\
\addlinespace[1pt]
\refsys{FullTraj} & 39.1 & 37.3 & 39.6 & 12.2 & 11.2 & 11.8 & 8.4 & 9.2 & 7.5 & 16.31 & 24.11 & 32.86 & \textbf{50.2} & \textbf{41.2} & \textbf{37.4} & \textbf{141.7} & \textbf{114.5} & \textbf{113.1} \\[-2pt]
 & \sd{0.8} & \sd{1.6} & \sd{1.7} & \ci{8.9--15.9} & \ci{7.8--14.9} & \ci{8.4--15.4} & \ci{5.3--11.9} & \ci{6.4--12.0} & \ci{5.1--10.0} & \sd{0.15} & \sd{1.46} & \sd{2.18} & \sd{1.3} & \sd{1.2} & \sd{3.8} & \sd{1.7} & \sd{6.0} & \sd{7.6} \\
\addlinespace[1pt]
A-Mem & 40.4 & 38.1 & 41.6 & 11.4 & 10.4 & \textbf{13.2} & 6.3 & 7.6 & 6.9 & 6.11 & 6.47 & 6.51 & 61.9 & 63.0 & 62.7 & 178.0 & 180.6 & 183.8 \\[-2pt]
 & \sd{2.4} & \sd{4.3} & \sd{1.6} & \ci{8.1--14.8} & \ci{7.1--13.9} & \ci{9.6--17.1} & \ci{4.0--9.0} & \ci{5.1--10.3} & \ci{4.6--9.6} & \sd{0.07} & \sd{0.36} & \sd{0.31} & \sd{1.6} & \sd{2.2} & \sd{1.7} & \sd{0.8} & \sd{5.0} & \sd{4.1} \\
\addlinespace[1pt]
Mem0 & 40.1 & 38.1 & 41.1 & \textbf{12.5} & 11.1 & 11.9 & 7.8 & 8.3 & 6.1 & 6.26 & 6.39 & \textbf{5.85} & 63.9 & 62.2 & 52.1 & 171.8 & 177.0 & 166.4 \\[-2pt]
 & \sd{2.2} & \sd{4.1} & \sd{1.1} & \ci{8.8--16.5} & \ci{7.8--14.6} & \ci{8.0--16.2} & \ci{5.2--10.6} & \ci{5.5--11.4} & \ci{3.4--9.2} & \sd{0.27} & \sd{0.33} & \sd{1.00} & \sd{1.6} & \sd{2.6} & \sd{9.8} & \sd{1.4} & \sd{4.2} & \sd{11.9} \\
\addlinespace[1pt]
MemOS & 36.8 & 37.3 & 35.3 & 10.7 & 11.3 & 9.9 & 9.2 & 9.3 & 9.9 & 6.49 & 6.56 & 6.68 & 63.4 & 65.8 & 66.5 & 189.5 & 188.0 & 189.2 \\[-2pt]
 & \sd{2.3} & \sd{2.3} & \sd{5.3} & \ci{7.7--13.9} & \ci{8.2--14.8} & \ci{7.3--12.7} & \ci{6.3--12.4} & \ci{6.3--12.4} & \ci{6.8--13.0} & \sd{0.53} & \sd{0.17} & \sd{0.43} & \sd{4.2} & \sd{2.6} & \sd{4.4} & \sd{5.1} & \sd{1.2} & \sd{5.8} \\
\addlinespace[1pt]
Supermemory & 40.6 & \textbf{40.6} & 39.3 & 11.7 & \textbf{13.1} & 11.2 & 6.4 & 7.8 & 7.2 & 6.39 & 6.25 & 6.20 & 62.8 & 61.6 & 60.7 & 183.8 & 182.3 & 183.3 \\[-2pt]
 & \sd{1.3} & \sd{2.7} & \sd{1.6} & \ci{8.5--15.2} & \ci{9.5--17.0} & \ci{7.8--15.0} & \ci{4.3--8.8} & \ci{5.3--10.5} & \ci{4.9--9.6} & \sd{0.14} & \sd{0.06} & \sd{0.06} & \sd{0.7} & \sd{1.6} & \sd{2.6} & \sd{0.6} & \sd{2.4} & \sd{2.5} \\
\addlinespace[1pt]
Claude-Mem & \textbf{41.6} & 40.1 & \textbf{42.6} & 12.2 & 11.5 & 12.1 & \textbf{5.9} & \textbf{6.8} & \textbf{4.8} & \textbf{5.82} & \textbf{6.03} & 6.23 & 57.9 & 59.8 & 60.1 & 180.0 & 181.1 & 180.8 \\[-2pt]
 & \sd{2.3} & \sd{3.8} & \sd{1.1} & \ci{8.9--15.7} & \ci{8.3--15.0} & \ci{8.9--15.5} & \ci{3.8--8.2} & \ci{4.2--9.7} & \ci{3.2--6.7} & \sd{0.09} & \sd{0.20} & \sd{0.41} & \sd{2.2} & \sd{1.4} & \sd{4.0} & \sd{2.6} & \sd{4.4} & \sd{4.5} \\
\bottomrule
\end{tabular}%
}
\end{table}

\textbf{Regressions offset gains, especially on already solvable tasks.}
For Improvement Tasks, Table~\ref{tab:improvement_results} shows that Resolved rises from 35.3\% for NoMem to 39.6\% with FullTraj and up to 42.6\% with memory systems. These gains are nevertheless limited by regressions: Rescue ranges from 9.9\% to 13.2\%, while Harm remains 4.8\%--9.9\%. Claude-Mem leads under Relevant and Repo history (41.6\% and 42.6\%) with low Harm (5.9\% and 4.8\%), while MemOS loses most of its gain to Harm.
Table~\ref{tab:preservation_results} shows that this issue is more pronounced on Preservation Tasks, where NoMem already resolves 94.0\% of tasks. Averaged over the five memory systems, Harm exceeds Rescue under Noisy (3.7\% vs.\ 3.2\%) and Repo history (3.2\% vs.\ 2.6\%), lowering Resolved below NoMem to 93.5\% and 93.3\%. Even modest regressions can thus erase the gains from memory.

\textbf{History noise tends to increase regressions rather than suppress Rescue.}
Table~\ref{tab:improvement_results} shows that, when averaged equally over the five structured memory systems, mean Rescue on Improvement Tasks remains nearly unchanged under Relevant, Noisy, and Repo history (11.7\%, 11.5\%, and 11.7\%). This indicates useful history remains exploitable despite unrelated history. 
From Relevant to Noisy history, mean Harm tends to rise (7.1\% to 8.0\%) and mean Resolved tends to fall (39.9\% to 38.8\%). Preservation Tasks in Table~\ref{tab:preservation_results} show the same trend: mean Rescue changes only slightly from 3.4\% to 3.2\%, whereas mean Harm increases from 2.5\% to 3.7\% and mean Resolved decreases from 94.9\% to 93.5\%.
FullTraj is more sensitive, with Harm rising from 0.7\% to 4.2\%. 
Thus, history noise tends to introduce regressions rather than reducing the benefit of useful history.

\textbf{Analysis of solving costs.}
Memory systems remain close to NoMem in solving cost, while FullTraj uses more input but fewer output tokens and steps.
The higher input is expected from including raw historical trajectories; Section~\ref{sec:results_fulltraj} analyzes the less obvious reductions in output and steps.

\begin{table}[t]
\centering
\caption{Results on 50 \textbf{Preservation Tasks}. Scenarios and
reporting conventions follow Table~\ref{tab:improvement_results}.}
\label{tab:preservation_results}
\scriptsize
\setlength{\tabcolsep}{1.6pt}
\renewcommand{\arraystretch}{0.95}
\resizebox{\linewidth}{!}{%
\begin{tabular}{l ccc @{\hskip 4pt} ccc @{\hskip 4pt} ccc @{\hskip 4pt} ccc @{\hskip 4pt} ccc @{\hskip 4pt} ccc}
\toprule
& \multicolumn{3}{c}{Resolved (\%)}
& \multicolumn{3}{c}{Rescue (\%)}
& \multicolumn{3}{c}{Harm (\%)}
& \multicolumn{3}{c}{Tok.\ In (M)}
& \multicolumn{3}{c}{Tok.\ Out (K)}
& \multicolumn{3}{c}{Steps} \\
\cmidrule(lr){2-4}\cmidrule(lr){5-7}\cmidrule(lr){8-10}\cmidrule(lr){11-13}\cmidrule(lr){14-16}\cmidrule(lr){17-19}
\textbf{System} & Rel. & Noi. & Repo & Rel. & Noi. & Repo & Rel. & Noi. & Repo & Rel. & Noi. & Repo & Rel. & Noi. & Repo & Rel. & Noi. & Repo \\
\midrule
\refsys{NoMem} & \multicolumn{3}{c}{94.0\,\sd{2.0}} & \multicolumn{3}{c}{--} & \multicolumn{3}{c}{--} & \multicolumn{3}{c}{3.08\,\sd{0.24}} & \multicolumn{3}{c}{34.9\,\sd{3.3}} & \multicolumn{3}{c}{130.5\,\sd{7.1}} \\
\addlinespace[1pt]
\refsys{FullTraj} & \textbf{97.3} & 92.7 & 90.0 & 4.0 & 2.9 & 2.7 & \textbf{0.7} & 4.2 & 6.7 & 7.22 & 11.80 & 22.99 & \textbf{29.6} & \textbf{24.4} & \textbf{22.5} & \textbf{95.1} & \textbf{83.4} & \textbf{81.2} \\[-2pt]
 & \sd{1.2} & \sd{2.3} & \sd{2.0} & \ci{0.7--8.0} & \ci{0.4--6.2} & \ci{0.7--5.3} & \ci{0.0--2.0} & \ci{1.6--7.3} & \ci{3.1--10.4} & \sd{0.43} & \sd{0.25} & \sd{1.03} & \sd{0.9} & \sd{3.1} & \sd{2.5} & \sd{5.6} & \sd{2.0} & \sd{3.6} \\
\addlinespace[1pt]
A-Mem & 95.3 & \textbf{94.0} & 93.3 & \textbf{4.4} & 3.1 & 2.7 & 3.1 & \textbf{3.1} & 3.3 & 3.30 & 3.50 & 3.55 & 37.8 & 41.6 & 39.1 & 128.9 & 131.2 & 134.8 \\[-2pt]
 & \sd{3.1} & \sd{2.0} & \sd{3.1} & \ci{1.1--8.9} & \ci{0.7--6.0} & \ci{0.7--5.3} & \ci{0.7--5.8} & \ci{0.9--5.8} & \ci{0.9--6.2} & \sd{0.34} & \sd{0.30} & \sd{0.13} & \sd{3.0} & \sd{1.3} & \sd{1.0} & \sd{5.8} & \sd{2.7} & \sd{2.3} \\
\addlinespace[1pt]
Mem0 & 96.7 & \textbf{94.0} & 91.3 & \textbf{4.4} & \textbf{4.0} & 1.6 & 1.8 & 4.0 & 4.2 & 3.47 & 3.37 & 3.07 & 41.3 & 39.4 & 35.9 & 128.2 & 127.8 & 122.6 \\[-2pt]
 & \sd{1.2} & \sd{3.5} & \sd{1.2} & \ci{0.9--8.9} & \ci{0.9--8.0} & \ci{0.0--3.6} & \ci{0.2--3.8} & \ci{1.1--8.0} & \ci{0.9--8.7} & \sd{0.35} & \sd{0.28} & \sd{0.23} & \sd{2.9} & \sd{1.1} & \sd{1.7} & \sd{5.3} & \sd{3.8} & \sd{2.4} \\
\addlinespace[1pt]
MemOS & 95.3 & 92.7 & 94.0 & 3.3 & 2.9 & 2.4 & 2.0 & 4.2 & 2.4 & 3.48 & \textbf{2.95} & 3.31 & 41.2 & 36.0 & 39.7 & 139.5 & 125.9 & 135.0 \\[-2pt]
 & \sd{2.3} & \sd{5.8} & \sd{0.0} & \ci{0.4--7.1} & \ci{0.4--6.2} & \ci{0.2--5.8} & \ci{0.2--4.2} & \ci{1.1--8.0} & \ci{0.4--5.1} & \sd{0.31} & \sd{0.26} & \sd{0.22} & \sd{1.0} & \sd{0.4} & \sd{2.8} & \sd{8.9} & \sd{5.2} & \sd{4.0} \\
\addlinespace[1pt]
Supermemory & 93.3 & 92.7 & 91.3 & 2.2 & 2.9 & 2.2 & 2.9 & 4.2 & 4.9 & \textbf{3.19} & 3.35 & 3.33 & 38.2 & 39.6 & 38.6 & 127.5 & 131.4 & 130.8 \\[-2pt]
 & \sd{1.2} & \sd{1.2} & \sd{3.1} & \ci{0.4--4.4} & \ci{0.7--5.8} & \ci{0.4--4.9} & \ci{0.7--5.6} & \ci{1.3--7.6} & \ci{1.8--8.7} & \sd{0.09} & \sd{0.29} & \sd{0.25} & \sd{0.7} & \sd{5.0} & \sd{1.0} & \sd{3.3} & \sd{4.5} & \sd{5.1} \\
\addlinespace[1pt]
Claude-Mem & 94.0 & \textbf{94.0} & \textbf{96.7} & 2.7 & 3.1 & \textbf{4.0} & 2.7 & \textbf{3.1} & \textbf{1.3} & \textbf{3.19} & 2.97 & \textbf{3.04} & 36.7 & 35.5 & 36.2 & 129.8 & 127.9 & 126.7 \\[-2pt]
 & \sd{2.0} & \sd{0.0} & \sd{1.2} & \ci{0.2--6.2} & \ci{0.7--6.5} & \ci{0.9--8.0} & \ci{0.4--5.6} & \ci{0.7--6.0} & \ci{0.0--3.1} & \sd{0.24} & \sd{0.18} & \sd{0.29} & \sd{3.5} & \sd{1.9} & \sd{4.0} & \sd{3.4} & \sd{3.4} & \sd{5.1} \\
\bottomrule
\end{tabular}%
}
\end{table}
\endgroup

\subsection{Attribution of Memory Gains}
\label{sec:results_rescue}

To understand the source of memory gains, we focus on the 68 Improvement Tasks that also fail in all three fresh NoMem runs. These tasks consistently fail without memory across independent runs, so improvements on these tasks can be more reliably attributed to memory.
Table~\ref{tab:repair_diagnostics} reports File, Line, and Edit Recall, which measure agreement with the reference solution patch at the file, edit-location, and edit-content levels, together with F2P and P2P for functional correctness and behavior preservation. Appendices~\ref{app:repair_metrics} and~\ref{app:patch_metrics} give the definitions.

\begin{table}[!htbp]
\centering
\caption{Task-solving outcomes and repair metrics on 68 tasks where all three NoMem runs fail.}
\label{tab:repair_diagnostics}
\scriptsize
\setlength{\tabcolsep}{1.6pt}
\renewcommand{\arraystretch}{0.95}
\resizebox{\linewidth}{!}{%
\begin{tabular}{
l
ccc @{\hskip 4pt}
ccc @{\hskip 4pt}
ccc @{\hskip 4pt}
ccc @{\hskip 4pt}
ccc @{\hskip 4pt}
ccc
}
\toprule
& \multicolumn{3}{c}{Resolved (\%)}
& \multicolumn{3}{c}{File Recall (\%)}
& \multicolumn{3}{c}{Line Recall (\%)}
& \multicolumn{3}{c}{Edit Recall (\%)}
& \multicolumn{3}{c}{F2P (\%)}
& \multicolumn{3}{c}{P2P (\%)} \\
\cmidrule(lr){2-4}
\cmidrule(lr){5-7}
\cmidrule(lr){8-10}
\cmidrule(lr){11-13}
\cmidrule(lr){14-16}
\cmidrule(lr){17-19}
\textbf{System}
& Rel. & Noi. & Repo
& Rel. & Noi. & Repo
& Rel. & Noi. & Repo
& Rel. & Noi. & Repo
& Rel. & Noi. & Repo
& Rel. & Noi. & Repo \\
\midrule

\refsys{NoMem}
& \multicolumn{3}{c}{0.00\,\sd{0.00}}
& \multicolumn{3}{c}{74.22\,\sd{0.37}}
& \multicolumn{3}{c}{18.41\,\sd{2.19}}
& \multicolumn{3}{c}{11.02\,\sd{0.58}}
& \multicolumn{3}{c}{18.10\,\sd{1.01}}
& \multicolumn{3}{c}{97.39\,\sd{0.22}} \\

\addlinespace[1pt]
\refsys{FullTraj}
& 11.27 & 12.75 & 11.27
& 71.61 & 70.59 & 70.96
& 22.16 & 20.59 & 19.64
& \textbf{14.77} & \textbf{13.90} & 12.09
& 26.10 & 25.81 & \textbf{27.53}
& 97.45 & 98.12 & 97.42 \\[-2pt]
& \sd{4.73} & \sd{1.70} & \sd{1.70}
& \sd{2.95} & \sd{2.73} & \sd{1.24}
& \sd{1.76} & \sd{1.05} & \sd{1.85}
& \sd{1.69} & \sd{1.64} & \sd{1.01}
& \sd{4.22} & \sd{0.19} & \sd{3.16}
& \sd{0.28} & \sd{0.72} & \sd{0.16} \\

\addlinespace[1pt]
A-Mem
& 10.29 & 10.78 & \textbf{12.75}
& 72.02 & \textbf{74.59} & 70.42
& 17.65 & 18.67 & 18.82
& 11.83 & 12.38 & 11.21
& 25.55 & 26.99 & 27.29
& 97.28 & 97.37 & 97.54 \\[-2pt]
& \sd{3.89} & \sd{4.25} & \sd{0.85}
& \sd{3.55} & \sd{2.09} & \sd{0.67}
& \sd{1.23} & \sd{1.70} & \sd{2.85}
& \sd{0.77} & \sd{0.83} & \sd{1.59}
& \sd{2.97} & \sd{3.55} & \sd{0.55}
& \sd{0.32} & \sd{0.29} & \sd{0.16} \\

\addlinespace[1pt]
Mem0
& \textbf{12.75} & 8.82 & 7.84
& 72.59 & 71.41 & \textbf{76.76}
& \textbf{22.54} & 18.21 & \textbf{22.05}
& 13.58 & 10.74 & \textbf{13.11}
& 26.74 & 23.93 & 25.45
& 97.64 & 97.54 & 97.28 \\[-2pt]
& \sd{0.85} & \sd{3.89} & \sd{1.70}
& \sd{4.22} & \sd{3.36} & \sd{1.84}
& \sd{1.20} & \sd{1.84} & \sd{1.07}
& \sd{1.18} & \sd{1.41} & \sd{1.52}
& \sd{2.15} & \sd{3.00} & \sd{1.54}
& \sd{0.03} & \sd{0.10} & \sd{0.00} \\

\addlinespace[1pt]
MemOS
& 10.29 & 12.25 & 8.82
& 72.55 & 72.30 & 72.22
& 19.06 & 20.85 & 18.77
& 12.76 & 13.06 & 11.98
& 25.90 & 27.04 & 23.15
& 97.57 & 97.05 & 97.51 \\[-2pt]
& \sd{1.47} & \sd{5.16} & \sd{7.35}
& \sd{1.00} & \sd{2.88} & \sd{0.50}
& \sd{1.39} & \sd{2.64} & \sd{2.26}
& \sd{1.39} & \sd{1.62} & \sd{1.27}
& \sd{1.01} & \sd{5.12} & \sd{6.43}
& \sd{0.17} & \sd{0.84} & \sd{0.40} \\

\addlinespace[1pt]
Supermemory
& 11.27 & \textbf{13.73} & 11.76
& 71.98 & 73.53 & 73.08
& 17.77 & \textbf{21.29} & 18.39
& 11.64 & 12.96 & 11.10
& \textbf{27.71} & \textbf{28.57} & 26.70
& \textbf{98.32} & \textbf{98.44} & \textbf{98.27} \\[-2pt]
& \sd{3.06} & \sd{6.63} & \sd{0.00}
& \sd{1.42} & \sd{0.77} & \sd{3.00}
& \sd{1.14} & \sd{3.60} & \sd{1.13}
& \sd{0.75} & \sd{2.67} & \sd{0.97}
& \sd{4.24} & \sd{5.67} & \sd{1.94}
& \sd{0.57} & \sd{1.06} & \sd{0.73} \\

\addlinespace[1pt]
Claude-Mem
& 11.76 & 9.31 & 8.82
& \textbf{72.92} & 71.85 & 72.02
& 20.03 & 19.73 & 17.69
& 12.45 & 11.29 & 9.69
& 27.65 & 24.45 & 23.79
& 97.39 & 97.49 & 97.60 \\[-2pt]
& \sd{4.41} & \sd{4.25} & \sd{1.47}
& \sd{1.39} & \sd{1.21} & \sd{0.60}
& \sd{1.57} & \sd{2.51} & \sd{1.68}
& \sd{2.06} & \sd{0.94} & \sd{0.80}
& \sd{3.10} & \sd{4.96} & \sd{2.05}
& \sd{0.16} & \sd{0.03} & \sd{0.11} \\

\bottomrule
\end{tabular}%
}
\end{table}

\textbf{Memory mainly improves fine-grained code editing.}
NoMem already achieves $74.22\%$ File Recall, and only 2 of the 18 non-NoMem settings exceed this baseline. In contrast, 13 settings exceed the $18.41\%$ Line Recall baseline and 16 exceed the $11.02\%$ Edit Recall baseline, with maximum gains of $4.13$ and $3.75$ pp, respectively. Memory thus rarely expands gold-file coverage but more consistently improves within-file localization and editing.

\textbf{Memory shows larger gains in functional behavior.}
F2P increases from $18.10\%$ for NoMem to $23.15\%$--$28.57\%$ across all 18 non-NoMem settings, while P2P remains high at $97.05\%$--$98.44\%$. Notably, the gains in functional metrics are much larger than those in gold-patch alignment metrics. This gap is expected because the gold patch represents only one valid solution, and a functionally correct solution need not reproduce its exact edits. 
Together with the Line and Edit Recall gains, the F2P improvements suggest that memory helps agents make more effective edits within relevant files, leading to better functional correctness.

\subsection{Attribution of Memory-Induced Regressions}
\label{sec:results_harm}
Section~\ref{sec:results_rescue} analyzes memory gains on tasks that NoMem consistently fails.
We examine the complementary case: 76 tasks solved in all three fresh NoMem runs.
Because failures can still arise from run-to-run variation, we attribute a failure to memory only when memory visibly shapes the failing decision or edit.
Manual inspection of failed trajectories from the five memory systems identifies 115 such cases.
In these trajectories, \textbf{agents tend to follow memory as a solution rather than reassess it against the current task context.} The resulting failures fall into four recurring, non-exclusive failure modes:
\begin{enumerate}[leftmargin=1.6em,labelsep=0.45em,itemsep=1.5pt plus 1pt,topsep=2.5pt plus 1pt,parsep=0pt]
    \item \textbf{Compatibility mismatch.} Agents adopt memory whose assumptions do not fit the current task.
    \item \textbf{Misapplied transfer.} Agents apply a historical solution to the wrong component or stage.
    \item \textbf{Scope expansion.} Agents extend a history-derived solution beyond what the current task needs.
    \item \textbf{Failure to revise.} Agents keep following memory-guided edits despite conflicting evidence.
\end{enumerate}

% Align the compact failure-mode chart with the analysis it supports.
\Needspace{11\baselineskip}
\begingroup
\setlength{\intextsep}{0pt}
\setlength{\columnsep}{14pt}
\begin{wrapfigure}{r}{0.40\textwidth}
    \centering
    \captionsetup{font=footnotesize,skip=4pt}
    \includegraphics[width=\linewidth,trim=0 5bp 0 4bp,clip]{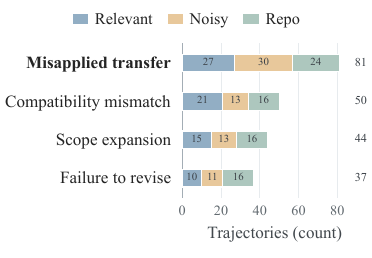}
    \caption[Counts of different failure modes.]{Counts of failure modes.}
    \label{fig:memory_failure_modes}
\end{wrapfigure}
\noindent\textbf{Failure-mode analysis across history scenarios.}
As shown in Figure~\ref{fig:memory_failure_modes}, misapplied transfer is the most common mode across all three history scenarios and is especially frequent under Noisy history. Trajectory analysis shows that it repeatedly occurs even after agents identify the correct files, indicating that adapting historical solutions to the current task remains difficult. 
Compatibility mismatch is most frequent under Relevant history, suggesting that relevance alone does not ensure reliable reuse because related history may contain incompatible assumptions. 
Scope expansion occurs at similar frequencies across scenarios, while failure to revise is least frequent but most common under Repo history, where more extensive history may increase reliance on memory-derived decisions and reduce revision when current-task feedback contradicts them.
\par
\endgroup

\subsection{How FullTraj Changes the Solving Process}
\label{sec:results_fulltraj}
To examine the reductions noted in Section~\ref{sec:results_overall}, we compare FullTraj with NoMem and the five memory systems across both subsets, examining where steps and action calls decrease.

\begin{figure}[htbp]
    \centering
    \includegraphics[width=\linewidth]{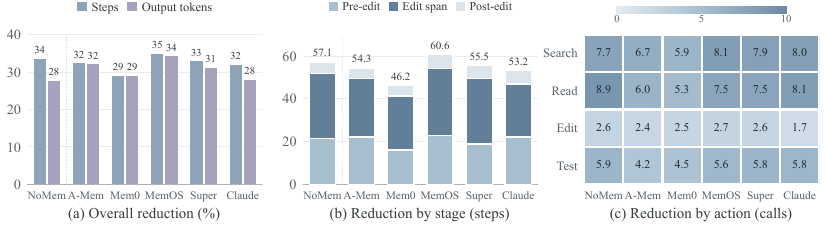}
    \caption{\textbf{FullTraj reductions in steps, output tokens, and action calls.}
    Comparisons with NoMem and five memory systems use pooled means over 183 instances.
    Super and Claude denote Supermemory and Claude-Mem.}
    \label{fig:fulltraj_behavior}
\end{figure}

\textbf{FullTraj substantially shortens the solving process.}
As shown in Figure~\ref{fig:fulltraj_behavior}a, FullTraj uses about 34\% fewer steps and 28\% fewer output tokens than NoMem, and 29\%--35\% fewer steps and 28\%--34\% fewer output tokens than the five memory systems.
All evaluated trajectories finish within the interaction budget, so the observed step reductions are not caused by budget truncation.
Figure~\ref{fig:fulltraj_behavior}b decomposes the 46.2--60.6-step reduction into pre-edit, edit-span, and post-edit stages, with the largest decrease occurring during the edit span (24.8--31.3 steps).
Figure~\ref{fig:fulltraj_behavior}c further shows that the shorter process involves less information gathering: FullTraj makes 5.9--8.1 fewer search and 5.3--8.9 fewer read calls per trajectory, with smaller reductions in test and edit calls.

\textbf{Understanding FullTraj's lower interaction cost.}
The five memory systems compress or select historical information and provide the resulting notes or tool summaries through retrieval or context injection.
See Appendix~\ref{app:memory_representations} for details.
This abstraction can lose information about the sequence of code inspection, failed attempts, revisions, and test outcomes, whereas FullTraj retains these events in execution order.
The results here show the potential benefit of preserving the historical solving process rather than only compressed summaries.
Meanwhile, Section~\ref{sec:results_harm} shows that historical solutions can mislead the agent when their assumptions do not hold for the current task.
Together, these results suggest that effective memory should preserve the solving process behind historical solutions while making their applicability explicit, a design implication discussed in Section~\ref{sec:failure_modes}.

\section{Design Directions}
\label{sec:failure_modes}

Our fine-grained analysis in Section~\ref{sec:results_rescue} shows that memory improves line- and edit-level recall more consistently than file-level recall. FullTraj further suggests that retaining more process information in execution order reduces the need to reconstruct details omitted by compressed memory representations commonly used by the evaluated memory systems, lowering interaction cost notably during editing; Section~\ref{sec:results_fulltraj} analyzes this effect. For code tasks, compressed memory is useful for high-level guidance but can be unreliable for fine-grained decisions: recovering omitted details requires extra search and read steps, while missing a critical condition can lead to incorrect localization or modification. Section~\ref{sec:results_harm} further shows that agents often continue following memory without sufficient reassessment, making such errors harder to correct. Together, these results suggest that memory should combine compact abstractions with finer process or code evidence for concrete edits, while agents should reassess historical guidance when current evidence no longer supports it.

Recent work has begun to address both sides. For memory representation and exposure, MCMA organizes experience across abstraction levels, MemHarness reconstructs retrieved experience for the current state, CoEvo-Mem co-adapts retrieval and memory through feedback, and CICL prioritizes evidence that can change the next action~\citep{liang2026mcma,wu2026memharness,ye2026coevomem,guan2026cicl}. These methods move beyond fixed memory replay, but still lack a mechanism to adaptively select historical evidence at the granularity required by each concrete decision. For memory reliance, MemHarness critiques experience before acting, BASM adds applicability boundaries and recovery guidance, MSCE associates skills with applicability, verification, and reliability signals, and Recuris uses working state and execution evidence for validation-gated memory evolution in a bounded recursive self-improvement (RSI) loop~\citep{wu2026memharness,lin2026basm,tang2026msce,yu2026recuris}. These methods increasingly control whether history should guide the agent, but this remains separate from deciding what granularity and amount of evidence to expose.

The remaining challenge is therefore to couple memory exposure and memory reliance at decision time. Building on multi-level memory, future systems should determine the granularity and amount of historical evidence needed for each decision. The selection of historical evidence should be coupled with how strongly that evidence influences the agent, and both should adapt as the current state and feedback evolve. More broadly, effective memory should serve as adaptive decision support rather than fixed historical guidance for the agent to follow.

\section{Conclusion and Limitations}
\label{sec:conclusion}
\textbf{Conclusion.}
Experiments on MemDev-Bench show that memory can improve task performance, but noisier histories tend to increase memory-induced regressions. Memory primarily improves within-file localization and fine-grained editing, while FullTraj shows the value of preserving process information, suggesting that applicability assessment and process preservation remain key challenges.

\textbf{Limitations.}
Our evaluation is limited to the SWE-bench Verified subset of MemDev-Bench under a single agent--model configuration. We also do not systematically compare our construction pipeline with existing methods. Broader evaluation and direct pipeline comparisons remain future work.

% Research statements and references.

\section*{Reproducibility Statement}
We provide detailed specifications to support reproduction of MemDev-Bench and our experiments.
Section~\ref{sec:mining} (\textit{Candidate Historical PR Mining}) describes the historical-task mining procedure, while Section~\ref{sec:anno} (\textit{Candidate Historical PR Annotation}) and Appendix~\ref{app:annotation} (\textit{Historical PR Relevance Annotation}) specify the relevance annotation procedure and complete annotation protocol.
Section~\ref{sec:task_construction} (\textit{Executable Historical Task Construction}) and Appendix~\ref{app:environment_construction} (\textit{Executable Historical Task Environment Construction}) describe executable task construction, environment preparation, and validation.
Section~\ref{sec:setup} (\textit{Experimental Setup}) and Appendix~\ref{app:experiment_setup} (\textit{Experimental Setup Details}) document the evaluation protocol, with agent and model configurations detailed in Appendix~\ref{app:agent_config} (\textit{Agent Configuration}) and information-isolation rules in Appendix~\ref{app:isolation} (\textit{Information Isolation Rules}).
Appendix~\ref{app:memory_representations} (\textit{Memory System Mechanisms}) describes the integration of the evaluated memory systems, and Appendix~\ref{app:metrics} (\textit{Metric Definitions}) defines all reported metrics and statistical procedures.

\section*{Ethics Statement}
Our benchmark is constructed from publicly available GitHub repositories, including issues, pull requests, commits, and code history, and is intended for research on code-agent evaluation. We use only information needed to construct and evaluate software-engineering tasks and do not intentionally collect sensitive personal information unrelated to the research objective. We preserve repository provenance and respect applicable licenses and terms. Human annotators assess only the technical relationships among development tasks. The study does not analyze annotators as research subjects or collect sensitive personal data from them.

\section*{AI Use Statement}
Generative AI tools were used to assist with literature retrieval and discovery, including identifying and summarizing relevant work; research ideation and execution, including experimental design, analysis, and presentation; and writing assistance, including translation, language editing, and improving clarity and concision. All AI-assisted outputs were reviewed and verified by the authors. The authors made the final decisions regarding the experiments, analysis, and presentation, and take full responsibility for the content of this work.

\section{References}
\printbibliography[heading=none]

\appendix
\section{Related Work}
\label{app:related_work}

\paragraph{Code agents and cross-task experience.}
Code agents such as SWE-agent~\citep{yang2024sweagent}, OpenHands~\citep{wang2025openhands}, and Claude Code~\citep{claude_code} combine repository navigation, code editing, and tool execution to address software development tasks.
Their interaction with the environment provides feedback for revising plans and code changes.
However, retaining context during one execution is different from maintaining reusable experience across tasks.
For evolving repositories, this experience may include code structure, implementation constraints, and test outcomes, whose usefulness depends on the current task and repository state.
Memory systems provide persistent representations and retrieval mechanisms for such reuse~\citep{AMem,Mem0,MemOS,SuperMemory,ClaudeMem}.
Our work studies how these systems affect performance on subsequent development tasks.

\paragraph{Memory systems for agents.}
Existing memory systems differ in what they retain and how experience is organized, updated, and exposed.
A-Mem~\citep{AMem} represents experience as structured notes with contextual attributes and links, and evolves existing notes as new memories arrive.
Mem0~\citep{Mem0} extracts and consolidates salient information into persistent memories, with an additional graph-based variant for relational information.
MemOS~\citep{MemOS} treats memory as a managed resource, using MemCubes to support organization and lifecycle operations across memory forms.
Supermemory~\citep{SuperMemory} combines source documents with extracted memories and graph relationships, supporting retrieval of both derived information and its underlying context.
Claude-Mem~\citep{ClaudeMem} captures tool-use observations and session summaries, and exposes them through search and chronological retrieval.
These approaches offer different balances between abstraction and retained execution detail.
Appendix~\ref{app:memory_representations} describes their integrations in our evaluation.

Recent methods increasingly learn memory management and condition reuse on current evidence.
MCMA learns to manage memories across abstraction levels~\citep{liang2026mcma}, while MemSkill learns and evolves reusable operations for extracting, consolidating, and pruning memories~\citep{zhang2026memskill}.
For retrieval, MemRL combines semantic matching with utility learned from execution feedback~\citep{zhang2026memrl}, and CoEvo-Mem jointly adapts the retrieval policy and memory bank~\citep{ye2026coevomem}.
CICL selects and compresses evidence according to its estimated influence on the next action~\citep{guan2026cicl}.
For applicability, MemHarness reconstructs retrieved experience for the current state~\citep{wu2026memharness}, while BASM records applicability boundaries, avoidance rules, and recovery guidance~\citep{lin2026basm}.
MSCE associates reusable skills with supporting evidence, applicability conditions, and verification signals~\citep{tang2026msce}.
Recuris uses working memory and execution evidence to guide localized, validation-gated updates to experiential memory~\citep{yu2026recuris}.
These advances motivate evaluating not only what is remembered, but also when historical guidance helps or misleads code agents.

\paragraph{Benchmarks for cross-task memory in development.}
Evaluating cross-task reuse requires earlier task experience to be available when an agent tackles a subsequent task.
SWE-bench and its Verified and Pro variants provide executable repository-level tasks, but their standard evaluation treats each task independently rather than specifying how memory accumulates across tasks~\citep{swebench,swebench_verified,swebench_pro}.
DreamBench-SWE~\citep{dreambench_swe} uses synthetic multi-session software tasks with executable oracles to test the retention and appropriate use of historical evidence.
ContextWeave~\citep{contextweave} reconstructs task streams from real longitudinal workflows, including coding, and evaluates memory from controlled workspace states with diagnostics for relevance and misleading recall.
These benchmarks address complementary aspects of cross-session memory but differ from benchmarks built from historical issue--PR tasks in versioned GitHub repositories.

Among repository-based benchmarks, SWE-ContextBench builds an experience pool from agent trajectories and evaluates retrieval on tasks linked through GitHub issues and PRs, including comparisons of memory frameworks~\citep{swecontextbench}.
Its reference-based construction permits tasks that share a resolving PR and does not explicitly enforce experience-before-target ordering in repository history.
SWE-Bench-CL~\citep{swebench_cl} constructs repository-specific sequences from SWE-bench Verified using timestamps and a difficulty curriculum, and records file-overlap dependencies.
Its proposed protocol resets each task to its own base commit while memory accumulates.
ChainSWE~\citep{chainswe} instead forms chains with overlapping code changes and carries agent-modified repository state across tasks, with additional comparisons of retained versus reset conversation history.
MemDev-Bench emphasizes human-verified, graded relevance of historical tasks to a base task and constructs Relevant, Noisy, and Repo histories with different compositions and lengths.
Historical tasks generate memory in chronological order, while each base-task evaluation starts from a fixed repository state with frozen memory.
This supports systematic comparisons of memory systems through task improvements, memory-induced regressions, fine-grained editing, and execution cost under a shared agent configuration.

Executable historical tasks also require reliable environment construction.
RepoLaunch in SWE-bench-Live and Repo2RLEnv automate repository environment setup~\citep{swebench_live,repo2rlenv}.
SWE-Factory and MEnvAgent coordinate setup, testing, and verification, including reuse of previously validated configurations~\citep{swefactory,menvagent}.
SWE-Universe further scales executable task construction through an agent with iterative self-verification~\citep{sweuniverse}.
Our pipeline focuses on repository-aware test--solution patch splitting and separate preparation of pre- and post-fix environments, and uses deterministic rules and validated experience before invoking an agent.
These mechanisms support the reconstruction and validation of historical tasks used for memory evaluation.
% ============================================================
% Appendix: Fine-grained Relation Types
% Table and figure are both anchored to this section.
% ============================================================
\section{Fine-grained Relation Types}
\label{app:fine-grained_relation_types}

On top of the three relevance levels Strong Related, Weak Related, and Unrelated, we define 20 fine-grained relation types: eight for Strong Related, six for Weak Related, and six for Unrelated. Each type describes the specific form of the relationship between a historical PR and a base task. This finer annotation separates the different kinds of help or distraction that historical information can provide, and lets us analyze how a memory system behaves under different types of history. Table~\ref{tab:fine_grained_relation_types} gives the complete definitions.

% ------------------------------------------------------------------ table --
\begin{table}[!tp]
\centering
\caption{\textbf{Fine-grained relation types between a historical PR and a base task.} Each relevance level has its own set of types, and an annotation must pick a type that belongs to the level it assigns.}
\label{tab:fine_grained_relation_types}
\footnotesize
\setlength{\tabcolsep}{0pt}
\renewcommand{\arraystretch}{1.33}
\begin{tabular}{@{}>{\raggedright\arraybackslash}p{0.215\linewidth}@{\hspace{9pt}}>{\raggedright\arraybackslash}p{0.755\linewidth}@{}}
\toprule
\textbf{Relation type} & \textbf{Definition} \\
\midrule
\addlinespace[1pt]
\multicolumn{2}{@{}l@{}}{\textbf{\textsc{Strong Related}}} \\
\addlinespace[2.5pt]
\emph{reusable implementation pattern}
& The historical PR contains an implementation pattern, code structure, algorithm, or repair strategy that can be reused directly, or applied to the base task after local adaptation. \\
\emph{same code path}
& The historical PR involves the same critical code path as the base task, or one directly connected to it, so its changes help localize where and how to fix it. \\
\emph{introducing change}
& The historical PR introduced or substantially modified the behavior, assumption, or implementation that later gave rise to the base task's problem, making it an important source for understanding the current issue. \\
\emph{same problem}
& The historical PR addresses essentially the same problem or requirement as the base task, possibly in a different version, component, or usage scenario. \\
\emph{same root cause}
& The surface symptoms of the historical PR and the base task differ, but the two share the same underlying technical cause, so the historical analysis or fix transfers directly to the current task. \\
\emph{design authority}
& The historical PR establishes a design decision, expected behavior, or system constraint that a correct implementation of the base task must respect. \\
\emph{same test surface}
& The historical PR contains tests that directly cover the behavior of the base task; its assertions and boundary conditions can guide the fix or its validation. \\
\emph{prerequisite implementation}
& The historical PR introduced functionality, an API, or infrastructure that the solution to the base task directly depends on or extends. \\
\addlinespace[3pt]
\midrule
\addlinespace[1pt]
\multicolumn{2}{@{}l@{}}{\textbf{\textsc{Weak Related}}} \\
\addlinespace[2.5pt]
\emph{same module}
& The historical PR and the base task involve the same module or component, which gives local implementation background, but its concrete changes do not transfer. \\
\emph{same file, different issue}
& The historical PR and the base task modify the same file but address different problems; its value lies mainly in conveying the structure or implementation context of that file. \\
\emph{same feature area}
& The historical PR and the base task belong to the same feature area but involve different sub-features, execution paths, or specific problems. \\
\emph{related API}
& The historical PR demonstrates the usage, semantics, or evolution of an API relevant to the base task, but does not provide the solution itself. \\
\emph{contextual precedent}
& The historical PR provides a design, implementation, or maintenance precedent that helps in understanding the codebase, but does not determine the specific fix. \\
\emph{adjacent test}
& The historical PR contains tests for adjacent or related behavior, giving a reference for testing style and expected behavior, without covering the base task. \\
\addlinespace[3pt]
\midrule
\addlinespace[1pt]
\multicolumn{2}{@{}l@{}}{\textbf{\textsc{Unrelated}}} \\
\addlinespace[2.5pt]
\emph{no overlap}
& The historical PR and the base task have no substantive relationship in functionality, implementation path, root cause, or design context. \\
\emph{different subsystem}
& The historical PR belongs to a different subsystem or component, and its changes provide no useful information for understanding or solving the current task. \\
\emph{coincidental file overlap}
& The historical PR and the base task modify the same file, but involve different code regions, functionality, or development goals, so the overlap is only surface-level. \\
\emph{dependency only}
& The historical PR and the base task share only a dependency, library, package, or infrastructure component, while the change itself carries no task-relevant information. \\
\emph{docs/CI only}
& The historical PR mainly concerns documentation, CI, formatting, release configuration, or other auxiliary changes, and offers no substantive help for the base task. \\
\emph{mass refactor}
& The historical PR touches code relevant to the base task only as a side effect of a large-scale refactor, migration, or repository-wide change. \\
\bottomrule
\end{tabular}
\end{table}

\begin{figure}[!t]
  \centering
  \includegraphics[width=0.98\linewidth]{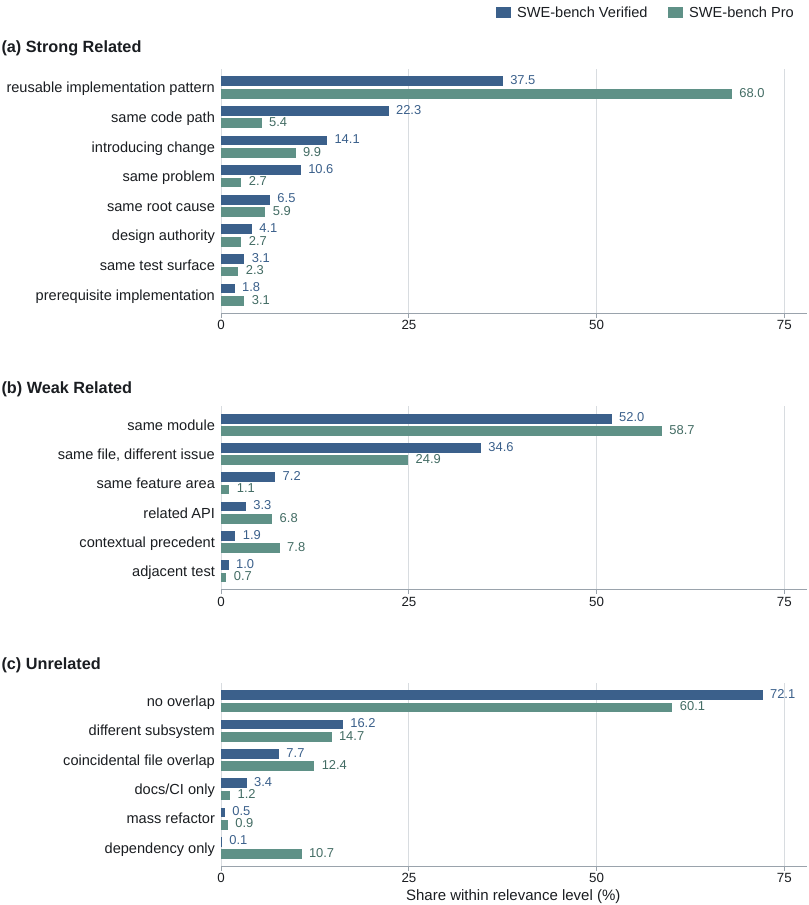}
  \caption{\textbf{Distribution of fine-grained relation types within each relevance level.} }
  \label{fig:fine_grained}
\end{figure}

Beyond the coarse relevance levels, these fine-grained types capture the
specific ways in which historical PRs can help, contextualize, or distract
from a base task. Strong Related types characterize directly transferable
implementation or design information, Weak Related types capture useful but
indirect context, and Unrelated types distinguish different forms of
non-useful or misleading overlap. Figure~\ref{fig:fine_grained} shows the
distribution of these relation types within each relevance level for the
Verified and Pro subsets.

% ----------------------------------------------------------------- figure --

% Keep this appendix's large floats before the following appendix heading.
\FloatBarrier

% ============================================================
% Appendix: Historical PR Relevance Annotation
% Requires \input{promptbox} in the preamble of main.tex.
% ============================================================
\section{Historical PR Relevance Annotation}
\label{app:annotation}

This appendix describes how candidate historical PRs are annotated for relevance in MemDev-Bench. It covers the annotation objective, the model instructions, and the input and output formats.

\subsection{Annotation Setup}

For each base task, we judge how much a candidate historical PR that precedes it can help a code agent solve the current task when used as historical development experience. Relevance here is not textual or topical similarity. It is whether the historical PR carries information of actual value for the current task. We use three levels:

\begin{itemize}[leftmargin=1.4em,labelsep=0.45em,itemsep=2.5pt,topsep=3pt,parsep=0pt]
\item \textbf{Strong Related.} The historical PR gives direct and actionable information, such as the key modification site, a reusable implementation, the underlying root cause, a design constraint, or directly relevant test behavior.
\item \textbf{Weak Related.} The historical PR gives clearly useful module, API, design, or test context, but not enough to determine the solution to the current task.
\item \textbf{Unrelated.} The historical PR gives no substantive help, even when the two overlap on the surface in keywords, files, or dependencies.
\end{itemize}

Beyond the level, the annotation model also selects one of the 20 fine-grained relation types defined in Table~\ref{tab:fine_grained_relation_types}. The selected type must belong to the selected level.

\subsection{System Prompt}

We reproduce the full system prompt below. It has four parts: the annotation objective and the three relevance levels (Box~\ref{box:sysprompt_levels}), the Strong relation types (Box~\ref{box:sysprompt_strong}), the Weak and Unrelated relation types (Box~\ref{box:sysprompt_weak_unrelated}), and the decision procedure, boundary rules, and output contract (Box~\ref{box:sysprompt_procedure}).

\begin{promptbox}[label={box:sysprompt_levels}]{System prompt (1/4): objective and relevance levels}
You are a relevance annotator for software engineering tasks in a code-agent memory benchmark. Given a \ptype{BASE task} that must be solved now and a \ptype{HISTORICAL PR} that occurred earlier, your job is to judge:

\promptask{If the code agent could read this HISTORICAL PR as historical development memory while solving the BASE task, what useful information would it obtain, and to what extent would that information help solve the current task?}

The judgment is not whether the two tasks are similar in text, topic, files, or keywords. It is whether the HISTORICAL PR contains historical development experience that helps the agent understand, localize, implement, or validate the BASE task.

You must make two related judgments:
\begin{penumerate}
\item Label the HISTORICAL PR as \ptype{Strong}, \ptype{Weak}, or \ptype{Unrelated}.
\item Select, from the relation types associated with that relevance level, the one that best describes the technical relationship between the BASE task and the HISTORICAL PR.
\end{penumerate}

\promptsec{Relevance levels}
\begin{pitemize}
\item \ptype{Strong}: the HISTORICAL PR provides direct and actionable information that lets the agent substantially narrow the solution space of the BASE task. Typically, the historical PR alone is already enough to say fairly concretely which code paths should be inspected or modified, which implementation approach to take, which root cause to focus on, which design constraints to follow, or how to construct and validate the relevant tests.
\item \ptype{Weak}: the HISTORICAL PR provides clear and valuable related context that helps the agent better understand the module, file, feature, API, design, or test behavior involved in the current task, but this information alone is not enough to determine how the current task should be implemented or fixed.
\item \ptype{Unrelated}: the HISTORICAL PR provides no substantive help for understanding or solving the BASE task, or the two overlap only superficially. Adding this PR to the agent context mainly introduces noise rather than useful historical development experience.
\end{pitemize}
\end{promptbox}

\begin{promptbox}[label={box:sysprompt_strong}]{System prompt (2/4): Strong relation types}
If the HISTORICAL PR can directly help solve the BASE task, select the most appropriate of the following:
\begin{pitemize}
\item \ptype{reusable implementation pattern}: the HISTORICAL PR contains an implementation pattern, code structure, algorithm, or repair strategy that can be reused directly, or applied to the BASE task after local adaptation.
\item \ptype{same code path}: the HISTORICAL PR involves the same critical code path as the BASE task, or one directly connected to it, and its changes directly help localize where to modify or determine how to implement the fix.
\item \ptype{introducing change}: the HISTORICAL PR introduced or substantially modified the behavior, assumption, or implementation that later gave rise to the problem addressed by the BASE task, and is therefore an important historical source for understanding the current issue.
\item \ptype{same problem}: the HISTORICAL PR addresses essentially the same problem or requirement as the BASE task without directly revealing its solution.
The problem may occur in a different version or usage scenario.
\item \ptype{same root cause}: the surface symptoms of the HISTORICAL PR and the BASE task may differ, but the two share the same underlying technical cause, so the historical analysis or fix transfers directly to the current task.
\item \ptype{design authority}: the HISTORICAL PR explicitly establishes, modifies, or explains a design decision, API contract, expected behavior, or system invariant that is directly relevant to the BASE task, and a correct implementation of the current task must respect these constraints.
\item \ptype{same test surface}: the tests in the HISTORICAL PR directly cover the behavior involved in the BASE task, and their inputs, assertions, boundary conditions, or expected behavior can directly guide the implementation or validation of the current task.
\item \ptype{prerequisite implementation}: the HISTORICAL PR introduced functionality, an API, infrastructure, or a key implementation that the solution to the BASE task directly depends on, calls, or extends.
\end{pitemize}
The core criterion for Strong is \ptype{actionability}. The following question can be used as a test:

\promptask{Based on this HISTORICAL PR alone, can I already tell the agent fairly concretely where to look or modify, and what kind of modification to make?}

If the answer is yes, Strong is normally the right choice.
\end{promptbox}

\begin{promptbox}[label={box:sysprompt_weak_unrelated}]{System prompt (3/4): Weak and Unrelated relation types}
\promptsectop{Weak relation types}
If the HISTORICAL PR cannot determine the solution to the BASE task but does provide clear and valuable related context, select from the following:
\begin{pitemize}
\item \ptype{same module}: the HISTORICAL PR and the BASE task involve the same module or component, which supplies local implementation background, but its concrete changes do not transfer directly to the current task.
\item \ptype{same file, different issue}: the HISTORICAL PR and the BASE task involve the same file but address different problems, functions, or code regions; the value lies mainly in understanding the structure or historical implementation of that file.
\item \ptype{same feature area}: the HISTORICAL PR and the BASE task belong to the same feature area but involve different sub-features, execution paths, or specific problems.
\item \ptype{related API}: the HISTORICAL PR demonstrates the usage, semantics, constraints, or evolution of an API relevant to the BASE task, but does not directly provide the solution to the current task.
\item \ptype{contextual precedent}: the HISTORICAL PR provides a design, implementation, or maintenance precedent related to the BASE task, which helps in understanding design intent or development background in the codebase, but does not determine how the current task should be modified.
\item \ptype{adjacent test}: the HISTORICAL PR contains tests for adjacent or conceptually related behavior, offering a reference for testing patterns, expected behavior, or boundary conditions, but does not directly cover the behavior involved in the BASE task.
\end{pitemize}
Weak must carry clear positive value. Coming from the same repository, occurring at a nearby time, sharing common keywords, or happening to modify the same frequently changed file is not sufficient to justify Weak.

\promptsec{Unrelated relation types}
If the HISTORICAL PR provides neither effective implementation information nor valuable context for the BASE task, select from the following:
\begin{pitemize}
\item \ptype{no overlap}: the HISTORICAL PR and the BASE task have no substantive relationship in functionality, implementation path, root cause, or design context that could help solve the current task.
\item \ptype{different subsystem}: the HISTORICAL PR belongs to a different subsystem or component, and its changes provide no useful information for understanding or solving the BASE task.
\item \ptype{coincidental file overlap}: the HISTORICAL PR and the BASE task modify the same file, but they involve different code regions, functionality, or development goals, so the file overlap is only a surface-level connection. Be especially careful with frequently modified files such as shared configuration files, shared utilities, fixtures, and \texttt{\_\_init\_\_.py}.
\item \ptype{dependency only}: the HISTORICAL PR and the BASE task share only a dependency, library, package, or infrastructure component, while the historical modification itself provides no implementation or design information relevant to the current task.
\item \ptype{docs/CI only}: the HISTORICAL PR mainly concerns documentation, CI, formatting, release configuration, licensing, or other auxiliary changes, and contains no code, design, or test information that would help implement or fix the BASE task.
\item \ptype{mass refactor}: the HISTORICAL PR touches files or code relevant to the BASE task only as a side effect of a large-scale refactor, rename, migration, formatting pass, import sweep, or repository-wide change, and contains no reusable task-relevant information.
\end{pitemize}
\end{promptbox}

\begin{promptbox}[label={box:sysprompt_procedure}]{System prompt (4/4): decision procedure, constraints, and output}
\promptsectop{Decision procedure}
Proceed in the following order:
\begin{penumerate}
\item First check whether the HISTORICAL PR provides information that can directly guide the solution of the BASE task, including but not limited to:
\begin{pitemize}
\item the files, functions, or critical code paths that the current task should inspect or modify;
\item an implementation approach that can be reused directly or adapted locally;
\item the underlying root cause of the current problem;
\item a design decision, API contract, or invariant that the current implementation must follow;
\item test logic, boundary conditions, or expected behavior that apply directly to the current task;
\item a historical implementation on which the current solution directly depends.
\end{pitemize}
If at least one such clear and actionable item exists, select \ptype{Strong}.
\item If the solution cannot be guided directly, check whether the HISTORICAL PR clearly improves the agent's understanding of the module, file, feature, API, design, or test behavior involved in the current task. If concrete and valuable context exists but is insufficient to determine the solution, select \ptype{Weak}.
\item If the historical PR neither guides the solution directly nor provides clear positive context, select \ptype{Unrelated}.
\end{penumerate}

\promptsec{Boundary cases}
Judge boundary cases conservatively:
\begin{pitemize}
\item If both Strong and Weak are defensible, but it cannot be shown that the historical information is sufficient to guide the current modification directly, select \ptype{Weak}.
\item If both Weak and Unrelated are defensible, but no concrete and valuable context added by the historical PR can be identified, select \ptype{Unrelated}.
\item If the evidence for relevance comes mainly from keywords, file names, the repository, temporal proximity, or other surface signals, and no actual technical relationship exists, select \ptype{Unrelated}.
\end{pitemize}

\promptsec{Important constraints}
\begin{pitemize}
\item Do not use information about the retrieval, mining, ranking, or recall method through which the candidate historical PR was obtained. Relevance must be judged entirely from the content of the BASE task and the HISTORICAL PR.
\item Do not assign Strong or Weak merely because similar wording appears in the title, body, problem statement, or commit messages.
\item Do not assume relevance merely because the two tasks come from the same repository.
\item Do not assume a useful relationship merely because the two tasks modified the same file. Inspect the code regions, functionality, and behavior actually involved.
\item The gold patch and the historical diff may be used to understand the real implementation relationship between the tasks, but do not judge by mechanical overlap of code lines; assess whether the changes convey development knowledge useful for the current task.
\item Weak is not the default category for uncertain samples. If no concrete positive contribution of the historical PR to the BASE task can be identified, select Unrelated.
\end{pitemize}

\promptsec{Confidence}
Set the confidence according to how clear the evidence behind the judgment is:
\begin{pitemize}
\item \ptype{high}: the evidence is direct, explicit, and essentially unambiguous, and independent reasonable annotators would normally reach the same conclusion.
\item \ptype{medium}: the evidence is sufficient, but the judgment requires combining several signals or applying the Strong/Weak or Weak/Unrelated boundary rules.
\item \ptype{low}: the judgment relies on weak evidence or substantial inference, and different reasonable annotators might disagree.
\end{pitemize}

\promptsec{Output}
Output a single JSON object and nothing else outside it. The JSON contains the following fields:
\begin{pitemize}
\item \texttt{label}: the relevance level, restricted to \texttt{strong}, \texttt{weak}, or \texttt{unrelated}.
\item \texttt{relation\_type}: one fine-grained relation type belonging to the selected relevance level.
\item \texttt{relationship}: one or two short sentences that objectively describe the concrete technical relationship between the BASE task and the HISTORICAL PR. Name specific files, functions, APIs, tests, issues, behaviors, or implementations wherever possible, rather than writing generic statements such as ``they are related to the same module.''
\item \texttt{confidence}: restricted to \texttt{high}, \texttt{medium}, or \texttt{low}.
\item \texttt{brief\_reason}: one to three short sentences explaining why the stated relationship implies the selected relevance level. For Strong and Weak, state what concrete help the historical PR offers the agent; for Unrelated, explain why the existing surface overlap does not amount to valuable historical experience.
\item \texttt{evidence\_spans}: one to three verbatim text or code spans from the BASE task or HISTORICAL PR input that directly support the judgment. Do not fabricate content that is absent from the input.
\end{pitemize}
\end{promptbox}

\subsection{User Message and Input Format}

We issue one independent model call per candidate historical PR. The user message adds no natural language task description. It provides the base task and the candidate historical PR as structured JSON:

\begin{jsonbox}{User message: input format}
{
  "base_task": {
    "problem_statement": "<problem statement>",
    "changed_files": ["<file 1>", "<file 2>"],
    "patch": "<gold patch>"
  },
  "historical_pr": {
    "title": "<PR title>",
    "body": "<PR body>",
    "changed_files": ["<file 1>", "<file 2>"],
    "diff_hunks": {
      "<path/to/file_1>": "<diff hunks>",
      "<path/to/file_2>": "<diff hunks>"
    },
    "commit_messages": ["<message 1>", "<message 2>"]
  }
}
\end{jsonbox}

This input gives the model three complementary signals.

\begin{itemize}[leftmargin=1.4em,labelsep=0.45em,itemsep=3pt,topsep=3pt,parsep=0pt]
\item \textbf{Development records.} The problem statement, the PR title and body, and the commit messages carry task intent, the problem description, and what the developers wrote down. They support relations that rest on the same problem, on design decisions, on explicit historical background, or on API semantics.
\item \textbf{Code structure.} The changed files on both sides show where the two tasks touch the repository, so the model can judge whether they involve the same module, file, or feature area. File overlap alone does not imply useful relevance, so this signal has to be read together with the concrete code changes.
\item \textbf{Code changes.} The gold patch of the base task and the diff hunks of the historical PR give the actual modifications, so the model compares implementation paths instead of surface wording. This matters most for Strong relations such as \emph{reusable implementation pattern}, \emph{same code path}, \emph{introducing change}, \emph{same root cause}, \emph{same test surface}, and \emph{prerequisite implementation}.
\end{itemize}

The annotation therefore combines development records, code structure, and actual code changes to decide whether historical development experience supplies reusable information for the current task.

We deliberately leave repository names, PR state, merge status, and GitHub labels out of the input. None of them carries evidence about the technical relationship between the two tasks. Merge status is a further risk: we also use it when we assemble the historical tasks, so exposing it to the model would offer a shortcut that bypasses the judgment we want to measure.

% Keep the subsection title and introduction with the short output schema.
\Needspace{18\baselineskip}
\subsection{Output Format}

For each base task--historical PR pair, the annotation model returns the following JSON:

\begin{jsonbox}{Model response: output format}
{
  "label": "strong | weak | unrelated",
  "relation_type": "<one fine-grained relation type>",
  "relationship": "<concrete technical relationship between the base task and the historical PR>",
  "confidence": "high | medium | low",
  "brief_reason": "<why this relationship implies the selected relevance level and what information the historical PR provides>",
  "evidence_spans": ["<evidence span 1>", "<evidence span 2>"]
}
\end{jsonbox}

The relation\_type must belong to the selected relevance level. For example, \emph{same root cause} can only accompany Strong, \emph{related API} can only accompany Weak, and \emph{coincidental file overlap} can only accompany Unrelated.

We keep relationship and brief\_reason as two separate fields. The first asks the model to describe the technical relationship that objectively holds between the two tasks. The second asks why that relationship implies the selected level and what information the historical PR supplies to the agent. This split stops the model from writing a generic justification for a label it has already committed to, and it lets human verifiers check relationship identification and relevance judgment separately.

The evidence\_spans field asks the model to quote the text or code that supports its judgment, so verifiers can quickly check whether the level and the relation type are grounded in the input.
\section{Executable Historical Task Environment Construction}
\label{app:environment_construction}

\subsection{Pipeline Overview}
\label{app:env_pipeline}

\begin{figure}[!htb]
    \centering
    \includegraphics[width=0.90\linewidth]{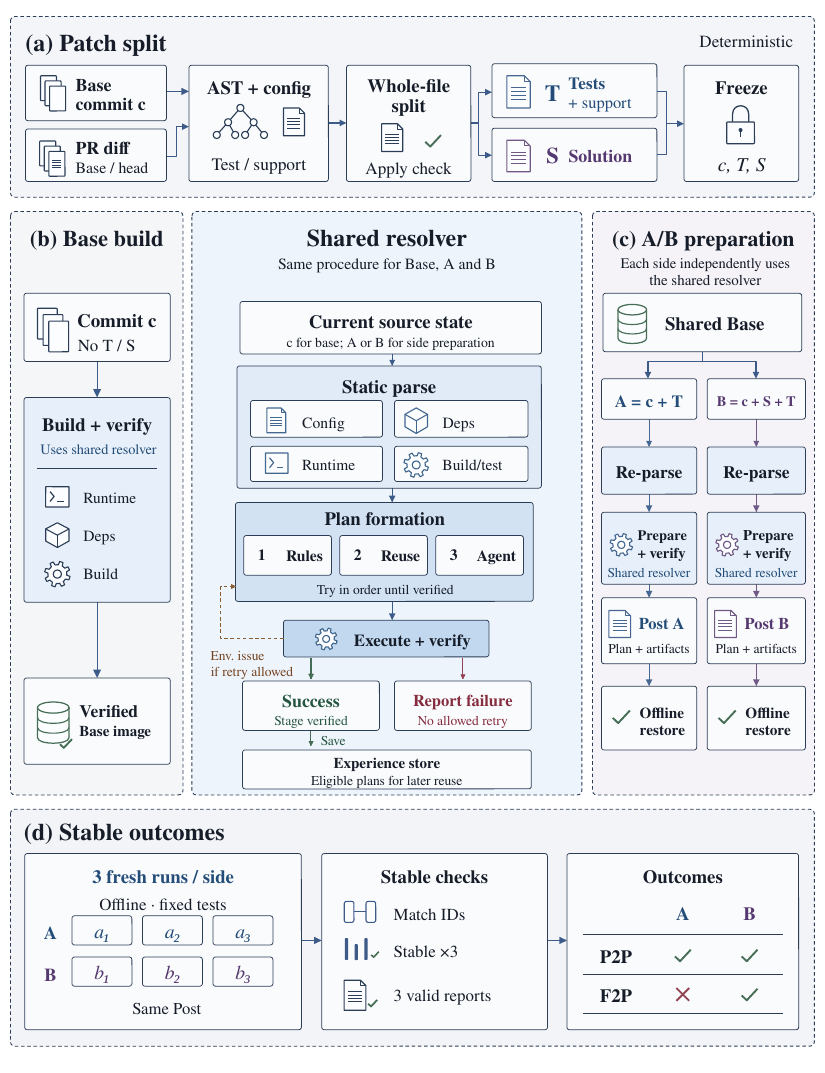}
    \caption{\textbf{Executable historical task construction pipeline.}
    (a) PR changes are split into test patch $T$ and solution patch $S$, and the task inputs are frozen.
    (b) The original commit $c$ is resolved and validated to construct a shared Base image.
    (c) The two patched source states independently invoke the shared resolver, and their plans and environment artifacts are frozen as Post A and Post B for offline restoration.
    (d) Stable outcomes are obtained from three fresh runs on each side.}
    \label{fig:construction_pipeline}
\end{figure}

As illustrated in Figure~\ref{fig:construction_pipeline}, we convert mined and relevance-annotated historical PRs into executable historical tasks.
Given the pre-PR source state and its code changes, the pipeline sequentially performs patch splitting, base-environment construction, side-specific environment preparation, and repeated testing, producing a frozen task definition, restorable execution environments, and per-test-case outcomes.
Base construction and side preparation share the same environment resolver, allowing a unified environment parsing, plan generation, and execution validation procedure to operate on both the original and patched source states.

\paragraph{Patch splitting and task freezing (Fig.~\ref{fig:construction_pipeline}a).}
Let $c$ denote the base commit before the PR, and let $T$ and $S$ denote the test and solution patches, respectively.
We classify PR changes based on repository test configuration, AST structures before and after the change, statically resolvable code relations, and task-role information.
Splitting is performed at the granularity of complete file patches, with tests and supporting changes required for task evaluation assigned to $T$ and implementation changes under evaluation assigned to $S$.
After checking patch applicability and composition order, we freeze $c,T,S$ and their provenance as fixed inputs to subsequent environment construction and testing.

\paragraph{Base-environment construction (Fig.~\ref{fig:construction_pipeline}b).}
We first invoke the shared resolver on the original commit $c$ to infer its runtime, dependencies, and build requirements, after which the corresponding environment plan is executed.
Neither $T$ nor $S$ is applied at this stage, ensuring that environment changes introduced by the PR are excluded from the base environment.
After validation, the resulting environment is frozen as the \textbf{Base image}, which provides the common starting point for side-specific preparation and subsequent testing.

\paragraph{Shared environment resolution and validation.}
The shared resolver operates on the source state available at each invocation. During Base construction, it analyzes the repository at the original commit $c$; during side-specific preparation, it analyzes the corresponding source state after the side's patch has been applied.
It first parses repository configuration, dependency declarations, runtime constraints, and build and test settings, then constructs an environment plan following the priority \textbf{Rules $\rightarrow$ Reuse $\rightarrow$ Agent}.
Deterministic rules first resolve configurations that can be derived directly from the repository. Previously validated solutions are reused only when their recorded applicability conditions match the current source state, and are then revalidated. The environment Agent completes any remaining unresolved parts of the plan.
All plans follow the same execution and validation procedure. Validated solutions and their applicability conditions are recorded for reuse in subsequent tasks.

\paragraph{Side-specific preparation and offline restoration (Fig.~\ref{fig:construction_pipeline}c).}
Starting from the same Base, we construct two source states:

$$
A=c+T,\qquad B=c+S+T,
$$

where addition denotes ordered patch application.
After the full patch sequence is applied, the shared resolver independently re-parses the environment requirements from the resulting source states $A$ and $B$.
Thus, dependency, runtime, or build changes introduced by $S$ affect only the preparation of $B$.
The validated environment plans and artifacts are then frozen as \textbf{Post A} and \textbf{Post B}.
Finally, we instantiate fresh containers from the Base image and restore each side using its frozen Post state, verifying that both environments can be reconstructed independently.

\paragraph{Repeated testing and stable outcomes (Fig.~\ref{fig:construction_pipeline}d).}
Both sides are evaluated through three independent runs.
Each run starts from a fresh offline container instantiated from the Base image, applies the corresponding patch, and restores the same frozen Post state for that side.
Runs share no writable state and do not reinstall dependencies.
We then align test outcomes between the two sides using the exact test-case identifiers recorded during execution.
A test case is considered stable only if all three runs contain complete records and produce the same outcome.
Cases that consistently fail on A and pass on B are labeled \textbf{F2P}, while those that consistently pass on both sides are labeled \textbf{P2P}.

\subsection{Comparison with Prior Work}
\label{app:env_comparison}

We compare prior historical-task construction methods along three dimensions: patch partitioning, patched-environment preparation and validation, and reuse of construction experience.

\subsubsection{Patch Partitioning with Repository and Task Context}
\label{app:env_patch_partitioning}

SWE-bench~\citep{swebench}, SWE-bench-Live~\citep{swebench_live}, and MEnvAgent~\citep{menvagent} identify test changes mainly from test-related keywords in file paths.
SWE-Factory~\citep{swefactory} further considers path tokens, naming conventions, and file types, while Repo2RLEnv~\citep{repo2rlenv} uses directory components, filename patterns, and exclusion rules.
Although these methods employ different heuristics, patch assignment still relies largely on file location and naming.
These signals alone cannot reliably determine whether a change defines the test criteria or implements the solution being evaluated.

Our partitioner combines repository-specific test configuration with AST-based analysis, statically resolvable code relations, and the role of each change in the task.
For example, a helper located under a test directory may also be used by non-test code.
If the task repairs this helper, the change should be treated as part of the solution; assigning it to the test patch would cause the pre-fix evaluation state to contain part of the intended repair.
Likewise, when the task itself is to correct an erroneous assertion, the assertion update is part of the solution rather than a test change shared across the pre- and post-fix states.
Repository-specific test configuration captures custom test roots and naming conventions that generic path-based rules may miss.
Our patch assignment therefore accounts for both repository-specific testing conventions and the functional role of each change in the task.

SWE-Universe~\citep{sweuniverse} also examines modification content, using an agent to distinguish test changes from solution changes.
In contrast, our partitioner combines repository configuration, code structure, statically resolvable usage relations, and task roles through deterministic rules, without invoking an agent during patch partitioning.

\subsubsection{Preparation and Validation of Patched Environments}
\label{app:env_patched_validation}

Existing systems already validate tasks across repository states before and after applying the reference patch.
SWE-bench~\citep{swebench} compares test behavior across these states and provides reusable environment configurations, with some configurations reinstalling the project after patching.
SWE-bench-Live~\citep{swebench_live} rebuilds both states, evaluating the pre-patch state once and the post-patch state three times in fresh containers.
Repo2RLEnv~\citep{repo2rlenv} compares individual test cases while executing both states within a shared sandbox.
SWE-Factory~\citep{swefactory} generates and repairs environment and evaluation scripts, while its outer F2P classifier relies on script-level exit codes.
MEnvAgent and SWE-Universe likewise use execution feedback to validate and adapt constructed environments~\citep{menvagent,sweuniverse}.
However, these methods do not always re-parse environment requirements from each complete patched source state.
As a result, patch-induced changes to runtimes, dependencies, or build requirements may be missed, causing an otherwise valid patch to fail under an environment that no longer matches its source state and confounding task validation with environment mismatch.

To address this mismatch, our design independently re-parses the runtime, dependency, build, and test requirements from each complete patched source state.
As described in Section~\ref{app:env_pipeline}, the resulting environment artifacts are frozen, restored in fresh offline containers, and reused across all three runs for that state.
This ties each observation to a specific source state and its prepared environment, while applying the same repetition criterion to both pre- and post-patch execution.

\subsubsection{Validated Reuse of Construction Experience}
\label{app:env_experience_reuse}
Prior work reuses construction knowledge in several forms.
SWE-bench reuses environment recipes and images~\citep{swebench}.
RepoLaunch, which supports SWE-bench-Live, can reuse historical environments and commands after execution checks~\citep{swebench_live}.
Repo2RLEnv can reuse cached bootstrap results without invoking its construction agent~\citep{repo2rlenv}.
SWE-Factory retrieves successful Dockerfiles and test scripts as references for model-based generation~\citep{swefactory}.
MEnvAgent retrieves and validates historical environments, repairing them when needed~\citep{menvagent}, while SWE-Universe trains its environment-construction model on successful construction trajectories~\citep{sweuniverse}.

Our pipeline integrates validated experience into the \textbf{Rules $\rightarrow$ Reuse $\rightarrow$ Agent} hierarchy described in Section~\ref{app:env_pipeline}.
Deterministic rules first resolve requirements that can be derived directly from the repository.
When unresolved requirements remain, the pipeline retrieves a previously validated construction plan with compatible applicability conditions, executes it in the current environment, and revalidates the result.
An agent is invoked only when neither rules nor reusable experience succeeds.
Successful agent-generated plans are then stored with their applicability conditions as executable experience for compatible future tasks.

Task-independent Base repairs that repeatedly succeed across independent tasks can further be incorporated into reusable parent-image capabilities after rebuilding and revalidation.
This allows construction experience to accumulate from task-level plans into stable Base capabilities, reducing repeated agent planning and exploratory execution while retaining validation on the current task.

\section{Experimental Setup Details}
\label{app:experiment_setup}

\subsection{Agent Configuration}
\label{app:agent_config}

We use Claude Code as the software-engineering agent and DeepSeek-V4-Flash as
the base model. To keep the six evaluation conditions comparable, the agent and
model configuration, the interaction budget, and the per-trial execution
settings are held fixed. Table~\ref{tab:agent_configuration} lists them.
FullTraj retains the complete trajectories generated by the same agent while solving the assigned historical tasks without memory, independent of task outcome.
NoMem results come from three fresh runs that are independent of the runs used to define the task subsets.
This avoids reusing the runs whose outcomes determine subset membership as the evaluation baseline.
Runs interrupted by infrastructure errors, such as API failures or container crashes, are rerun under the same configuration, and the same rule applies to every condition.

\begin{table}[!htbp]
\centering
\caption{Shared agent and execution configuration. Every setting is identical
across all evaluated systems and all six evaluation conditions.}
\label{tab:agent_configuration}
\small
\setlength{\tabcolsep}{0pt}
\renewcommand{\arraystretch}{1.2}
\begin{tabular}{@{}
>{\raggedright\arraybackslash}p{0.255\linewidth}
@{\hspace{0.03\linewidth}}
>{\raggedright\arraybackslash}p{0.715\linewidth}@{}}
\toprule
\textbf{Setting} & \textbf{Configuration} \\
\midrule
\multicolumn{2}{@{}l@{}}{\emph{Agent and base model}} \\
\addlinespace[1.5pt]
Agent & Claude Code 2.1.170 \\
Base model & DeepSeek-V4-Flash \\
Model routing & The primary route, all Claude Code model aliases, and the
subagent route use DeepSeek-V4-Flash \\
Reasoning effort & High \\
Interaction budget & At most 200 turns per task \\
\addlinespace[4pt]
\multicolumn{2}{@{}l@{}}{\emph{Execution environment}} \\
\addlinespace[1.5pt]
Evaluation framework & Harbor 0.3.0~\citep{harbor} \\
Task environment & Docker containers \\
Resources & 1 CPU, 4~GB memory, and 10~GB storage per trial \\
Time limits & 3{,}000~s for the agent, 3{,}000~s for verification, and
3{,}600~s for each environment-start attempt \\
\bottomrule
\end{tabular}
\end{table}

\subsection{Information Isolation Rules}
\label{app:isolation}

We apply a unified closed-book information isolation protocol while the agent
solves both historical tasks and the base task.
The protocol prevents access to task solutions and
repository history while preserving the local tools, dependency installation,
and task-required connectivity needed for normal software development.
The agent can use the historical trajectories provided for the current instance and information retrieved from its designated memory store. 
The memory store is read-only while the agent solves the base task.
Each base task uses an independent memory store, so memory never carries across base tasks.
Claude Code's built-in persistent memory is disabled in all conditions.

\subsubsection{Closed-Book System Prompt}
\label{app:closed_book_prompt}

During both historical- and base-task solving, we append the following policy to Claude Code's system prompt.

\begin{promptbox}[label={box:closed_book_prompt}]{Closed-book system prompt}

You are participating in a strict closed-book software-engineering benchmark.

\promptsectop{PERMITTED EVIDENCE}

Your primary obligation is to solve the task using your own reasoning. You may
use only:
\begin{penumerate}
\item The task instruction provided to you.
\item Files in the current working-tree checkout.
\item Output produced by local builds, tests, linters, type checkers, debuggers,
and reproduction scripts.
\item Task-provided runtime services that are necessary to reproduce or test the
requested behavior.
\item Historical trajectories supplied for the current instance and information
retrieved from its designated memory store, when available.
\end{penumerate}
Use only the historical information provided for the current task.
You may read the supplied trajectories and query the designated memory store through its provided interface, including network access when required.
The restrictions below on external retrieval and repository history do not prohibit this access.
During base-task solving, the history is frozen and the memory store is read-only.

\promptsec{NETWORK POLICY}

Normal network connectivity may be available because some tasks require
external services or installation of missing third-party dependencies. Network
access is an execution transport only; it must never be used as an information,
research, retrieval, or answer source.

You may use the network only to:
\begin{pitemize}
\item Install a missing third-party build, runtime, or test dependency.
\item Communicate with an endpoint explicitly required to reproduce or test the
task.
\item Exercise HTTP, API, OAuth, webhook, proxy, package-manager, or other
network behavior required by the task.
\end{pitemize}

You must not use the network to:
\begin{pitemize}
\item Search for the task, issue, error message, expected patch, or solution.
\item Read documentation, tutorials, blog posts, Q\&A pages, release notes,
changelogs, or external examples to help solve the task.
\item Access search engines, public code hosts, issue trackers, pull requests,
commits, patches, benchmark records, leaderboards, or hidden tests.
\item Download, install, or inspect another copy or version of the target
project.
\item Query another model, assistant, search API, retrieval service, or external
knowledge service.
\item Use curl, wget, Python, package managers, Git, MCP tools, subagents,
browsers, or any other mechanism to bypass these restrictions.
\end{pitemize}

Do not use WebSearch or WebFetch. If the task instruction contains a URL, issue
number, pull request number, or commit reference, treat it only as text supplied
by the task. Do not open or retrieve it.

\promptsec{SOURCE AND GIT POLICY}

Work only with the current checkout.

The only permitted Git commands are:
\begin{pitemize}
\item git status
\item git diff limited to the current working tree or index
\item git diff HEAD limited to the current working tree or index
\end{pitemize}

Do not use any other Git command, including log, show, blame, reflog, rev-list,
cat-file, fetch, pull, clone, checkout, switch, restore, reset, cherry-pick,
merge, rebase, archive, branch, tag, or worktree.

Do not inspect .git directly. Do not access Git objects, refs, logs,
alternate object databases, remote references, stashes, or historical
revisions. Do not reconstruct historical content by any indirect method.

Do not read the target project from site-packages, dist-packages,
package-manager caches, another checkout, an installed wheel, a source archive,
or any location outside the current checkout. You may inspect genuinely
separate third-party dependencies when necessary to understand their local
interfaces or runtime behavior.

\promptsec{BENCHMARK PRIVACY POLICY}

Do not access benchmark-private /tests or /solution paths,
gold patches, verifier implementation details, hidden tests, expected outputs,
grading metadata, or results from other agents.

Repository-owned tests inside the current checkout are allowed. You may add a
focused reproduction test when useful, but do not delete, skip, weaken,
rewrite, or manipulate tests merely to make them pass.

\promptsec{TOOL POLICY}

You may use local file-reading, file-editing, shell, build, test, debugging, and
static-analysis tools. Subagents, if available, must follow exactly the same
closed-book restrictions and may use only the same permitted evidence.

Do not use a permitted tool to simulate or replace a prohibited tool. Do not
encode, disguise, split, delegate, or indirectly execute a prohibited
operation.

\promptsec{WORKING METHOD}

Investigate the current code, reproduce the problem locally when possible,
identify the root cause, implement the smallest correct fix, and validate it
with relevant local tests.

Base conclusions on the permitted evidence.
If information is missing, reason from that evidence rather than using prohibited external sources.

If an operation is blocked or prohibited, do not retry it through another
command, tool, process, subagent, encoding, or indirect path. Continue solving
from the permitted evidence.

At completion, provide a concise summary of the completed task and the work
performed.

Solve the task yourself. Do not obtain the answer, implementation, patch, or
decisive technical guidance from any source outside the permitted evidence.

\end{promptbox}

\subsubsection{Tool-Level Enforcement}
\label{app:closed_book_tools}

The system prompt states the closed-book information boundary; two layers of
programmatic control enforce it during execution. First, the Claude Code CLI
turns off the tools whose only purpose is external retrieval or work outside
the current checkout. Second, a PreToolUse guard validates every request to the
remaining tools before execution and rejects commands or arguments that would
violate the protocol. Table~\ref{tab:closed_book_tool_coverage} summarizes tool
availability under this scheme.
The supplied trajectories and designated memory interface remain accessible under both enforcement layers.

\begin{table}[!htbp]
\centering
\caption{Tool availability under the information-isolation protocol. Tools that
remain available are checked request by request rather than turned off, so
normal local development is preserved.}
\label{tab:closed_book_tool_coverage}
\small
\setlength{\tabcolsep}{0pt}
\renewcommand{\arraystretch}{1.2}
\begin{tabular}{@{}
>{\raggedright\arraybackslash}p{0.235\linewidth}
@{\hspace{0.025\linewidth}}
>{\raggedright\arraybackslash}p{0.125\linewidth}
@{\hspace{0.025\linewidth}}
>{\raggedright\arraybackslash}p{0.56\linewidth}@{}}
\toprule
\textbf{Tool category} & \textbf{Availability} & \textbf{Enforcement} \\
\midrule
Bash
& Available
& The guard validates each request before execution and inspects the complete
command string. \\
\addlinespace[2pt]
Read, Write, Edit, NotebookEdit, Glob, Grep
& Available
& The guard validates each request for protected paths, Git metadata, and
alternate copies of the target project. \\
\addlinespace[2pt]
WebSearch, WebFetch
& Disabled
& Turned off in the CLI and additionally rejected by the guard if invoked,
removing the direct route to external information. \\
\addlinespace[2pt]
EnterWorktree, ExitWorktree
& Disabled
& Turned off in the CLI and additionally rejected by the guard if invoked,
keeping the agent inside the current checkout. \\
\bottomrule
\end{tabular}
\end{table}

For the tools that remain available, enforcement happens at the request level
rather than by turning the tool off. The guard inspects commands, paths, and
relevant arguments, and blocks access to external solutions, repository
history, benchmark-private information, and alternate copies of the target
project, while leaving the operations needed for normal task solving intact.
Table~\ref{tab:closed_book_guard_checks} lists the main rule families and the
behavior each one preserves.

\begingroup
\footnotesize
\setlength{\tabcolsep}{0pt}
\renewcommand{\arraystretch}{1.18}
\setlength{\LTpre}{9pt}
\setlength{\LTpost}{9pt}
\setlength{\LTcapwidth}{\textwidth}
\begin{longtable}{@{}
>{\raggedright\arraybackslash}p{0.155\linewidth}
@{\hspace{0.025\linewidth}}
>{\raggedright\arraybackslash}p{0.465\linewidth}
@{\hspace{0.025\linewidth}}
>{\raggedright\arraybackslash}p{0.30\linewidth}@{}}
\caption{Main request-level checks enforced by the PreToolUse guard. Each rule
family blocks a route to external answers or repository history while keeping
the corresponding development operation available.}
\label{tab:closed_book_guard_checks}\\
\toprule
\textbf{Rule family} & \textbf{Restricted behavior} & \textbf{Preserved behavior} \\
\midrule
\endfirsthead
\multicolumn{3}{@{}l}{\tablename~\thetable{} continued from the previous page}\\
\toprule
\textbf{Rule family} & \textbf{Restricted behavior} & \textbf{Preserved behavior} \\
\midrule
\endhead
\midrule
\multicolumn{3}{r@{}}{\emph{Continued on the next page}}\\
\endfoot
\bottomrule
\endlastfoot

Target-project copies
& Inspecting the target project under \texttt{site-packages} or
\texttt{dist-packages}, package-metadata queries that locate such copies, and
imports of the target project from outside the current checkout.
& Genuinely separate third-party dependencies may still be inspected. \\
\addlinespace[2pt]

Shell and package acquisition
& Remote-access and source-acquisition tools (\texttt{gh}, \texttt{ssh},
\texttt{scp}, \texttt{sftp}, \texttt{ftp}, \texttt{netcat}, \texttt{rsync});
\texttt{pip download}/\texttt{wheel}; \texttt{-{}-target}/\texttt{-t}
installation; and URL- or VCS-based installation.
& Ordinary installation of named third-party dependencies. \\
\addlinespace[2pt]

Git and metadata
& Git history, any Git operation other than the permitted commands, direct
access to \texttt{.git}, Git plumbing utilities, and Git-metadata environment
variables.
& \texttt{git status} and \texttt{git diff} on the current working tree or
index, optionally against \texttt{HEAD}. \\
\addlinespace[2pt]

Benchmark-private paths
& Access to the benchmark-private \texttt{/tests} and \texttt{/solution}
paths.
& Repository-owned source files and tests inside the current checkout. \\
\addlinespace[2pt]

Explicit network access
& Explicit URLs in \texttt{curl}, \texttt{wget}, and recognized inline-Python
calls (\texttt{requests}, \texttt{httpx}, \texttt{aiohttp}, \texttt{urllib},
\texttt{http.client}, \texttt{socket}) that target code hosts (GitHub and
\texttt{githubusercontent.com}, GitLab, Bitbucket), Q\&A sites (Stack Overflow
and Stack Exchange), search engines (Google, Bing, DuckDuckGo, Brave Search),
or SWE-bench.
& Networking needed for ordinary dependency installation and task execution. \\
\addlinespace[2pt]

Guard bypass
& Attempts to unset or override \texttt{LD\_PRELOAD} or network-guard endpoint
variables, or to bypass the execution guard by any other means.
& Requests that satisfy the isolation checks proceed. \\
\addlinespace[2pt]

File and search requests
& Read, Write, Edit, NotebookEdit, Glob, and Grep requests that touch Git
metadata, benchmark-private paths, or alternate installed copies of the target
project.
& Requests confined to permitted files in the current checkout. \\

\end{longtable}
\endgroup

Overall, the protocol implements a closed-book environment rather than a fully
offline sandbox: the connectivity required for dependency installation and task
execution is preserved, while access to external information and repository
history is restricted programmatically.
\section{Broad Evaluation of Memory Systems}
\label{app:all_systems}

Section~\ref{sec:setup} studies five representative memory systems across all
six evaluation conditions. To cover a wider range of memory-system designs at a
manageable evaluation cost, we further evaluate 29 additional systems under the
Relevant History scenario on a fixed subset of 150 base tasks. The subset draws
on both Improvement Tasks and Preservation Tasks so that the comparison covers
memory-induced gains and potential regressions while keeping the evaluation
scale bounded, comprising 120 of the 133 Improvement Tasks and 30 of the 50
Preservation Tasks. The same 150 tasks are used for every system.
Table~\ref{tab:broad_memory_results} reports the complete results, with NoMem
included as a reference baseline. To control the cost of this large-scale
evaluation, each of the 29 systems is evaluated with a single run, whereas the
NoMem baseline is averaged over three runs and reported as mean and sample
standard deviation.

The remaining metrics follow the definitions in Appendix~\ref{app:metrics}.
Resolved and File Recall are task-level rates, F2P and P2P are micro-averaged
over their corresponding test cases, and the Tokens and Steps columns report
per-task means over the same tasks. Systems are ranked primarily by Resolved,
which measures final task-solving success. Ties are broken in turn by F2P, File
Recall, and P2P: F2P captures partial repair progress on issue-specific tests,
File Recall reflects whether the agent reaches the reference files, and P2P
measures preservation of pre-existing passing behavior. 
If two systems tie on all four performance metrics, we prefer the one with lower solving cost. We compare input tokens, output tokens, and agent steps in that order.

\begingroup
\footnotesize
\setlength{\tabcolsep}{2.5pt}
\renewcommand{\arraystretch}{1.15}
\setlength{\LTpre}{10pt}
\setlength{\LTpost}{6pt}
\setlength{\LTcapwidth}{\textwidth}

\begin{longtable}{@{}>{\raggedright\arraybackslash\hspace{0pt}}p{0.450\textwidth}rrrrrrr@{}}

\caption{Evaluation of 29 memory systems under the Relevant History scenario on the same 150 base tasks.}
\label{tab:broad_memory_results}\\

\toprule
& \multicolumn{4}{c}{\textbf{Task performance}}
& \multicolumn{3}{c}{\textbf{Solving cost}} \\
\cmidrule(lr){2-5}\cmidrule(lr){6-8}
\textbf{Memory system}
& \textbf{Resolved} & \shortstack{\textbf{File}\\\textbf{Recall}} & \textbf{F2P} & \textbf{P2P}
& \multicolumn{2}{c}{\textbf{Tokens}} & \textbf{Steps} \\
\cmidrule(lr){6-7}
& (\%) & (\%) & (\%) & (\%) & In (M) & Out (K) & \\
\midrule
\endfirsthead

\multicolumn{8}{@{}l}{\tablename~\thetable{} continued from the previous page}\\
\toprule
& \multicolumn{4}{c}{\textbf{Task performance}}
& \multicolumn{3}{c}{\textbf{Solving cost}} \\
\cmidrule(lr){2-5}\cmidrule(lr){6-8}
\textbf{Memory system}
& \textbf{Resolved} & \shortstack{\textbf{File}\\\textbf{Recall}} & \textbf{F2P} & \textbf{P2P}
& \multicolumn{2}{c}{\textbf{Tokens}} & \textbf{Steps} \\
\cmidrule(lr){6-7}
& (\%) & (\%) & (\%) & (\%) & In (M) & Out (K) & \\
\midrule
\endhead

\midrule
\multicolumn{8}{r@{}}{\emph{Continued on the next page}}\\
\endfoot

\bottomrule
\endlastfoot

causal-memory~\citep{CausalMemory}
& 52.67 & 81.56 & 79.84 & 99.47 & 6.18 & 56.50 & 186.35 \\

\addlinespace[2.5pt]

ChronoHybridMem~\citep{ChronoHybridMem}
& 52.00 & 83.34 & 81.29 & 99.42 & 6.62 & 59.31 & 186.68 \\

Memory Bear~\citep{MemoryBear}
& 52.00 & 79.90 & 80.50 & 99.46 & 6.62 & 55.95 & 180.39 \\

\addlinespace[2.5pt]

TencentDB Agent Memory~\citep{TencentDBAgentMemory}
& 51.33 & 82.23 & 80.76 & 99.45 & 5.86 & 55.94 & 184.69 \\

TraceMem~\citep{TraceMemChallenge}
& 51.33 & 81.40 & 80.63 & 99.44 & 6.52 & 56.16 & 179.07 \\

LightMem v2.0~\citep{LightMem}
& 51.33 & 82.79 & 80.50 & 99.44 & 5.44 & 51.16 & 180.77 \\

picoAgent~\citep{PicoAgent}
& 51.33 & 81.73 & 80.24 & 99.44 & 5.91 & 53.42 & 172.89 \\

memoria~\citep{MemoriaFairme}
& 51.33 & 83.51 & 79.97 & 99.44 & 6.46 & 56.57 & 186.88 \\

Hindsight~\citep{Hindsight}
& 51.33 & 81.87 & 79.91 & 99.44 & 6.52 & 54.13 & 178.22 \\

\addlinespace[2.5pt]

Cognee~\citep{Cognee}
& 50.67 & 84.12 & 80.11 & 99.42 & 6.54 & 58.77 & 189.65 \\

Aranea-Agents~\citep{AraneaAgentsMemory}
& 50.67 & 83.20 & 80.11 & 99.25 & 6.55 & 54.73 & 181.82 \\

Infini Memory~\citep{InfiniMemory}
& 50.67 & 82.84 & 80.11 & 99.42 & 6.32 & 59.09 & 185.58 \\

\addlinespace[2.5pt]

SQLite FTS Baseline~\citep{SQLiteFTSBaseline}
& 50.00 & 81.81 & 80.37 & 99.43 & 5.29 & 50.85 & 176.49 \\

ReFind~\citep{ReFind}
& 50.00 & 81.67 & 80.37 & 99.43 & 6.74 & 58.85 & 194.23 \\

LightMem~\citep{LightMem}
& 50.00 & 83.90 & 79.71 & 99.45 & 5.87 & 54.84 & 182.04 \\

\addlinespace[2.5pt]

MemPalace~\citep{MemPalace}
& 49.33 & 83.62 & 79.97 & 99.46 & 7.04 & 58.91 & 189.95 \\

TencentDB Agent Memory (4o-mini)~\citep{TencentDBAgentMemory}
& 49.33 & 83.09 & 79.71 & 99.43 & 5.81 & 55.30 & 185.96 \\

\addlinespace[2.5pt]

EverMemOS~\citep{EverMemOS}
& 48.67 & 80.79 & 79.45 & 99.46 & 6.95 & 59.42 & 185.35 \\

\addlinespace[2.5pt]

Chronicle Memory~\citep{ChronicleMemory}
& 48.00 & 80.51 & 79.84 & 99.44 & 5.84 & 54.11 & 182.47 \\

openProustian Agent Memory~\citep{OpenProustianAgentMemory}
& 48.00 & 81.95 & 79.45 & 99.41 & 6.36 & 56.11 & 180.93 \\

RecMem~\citep{RecMem}
& 48.00 & 80.40 & 79.31 & 99.44 & 7.01 & 62.34 & 194.99 \\

SimpleMem~\citep{SimpleMem}
& 48.00 & 82.34 & 79.05 & 99.47 & 6.53 & 54.72 & 176.94 \\

\addlinespace[2pt]
\midrule
\textbf{NoMem (3 runs)}
& \shortstack{\textbf{47.78}\\[-1pt]{\scriptsize$\mathbf{\pm 1.54}$}}
& \shortstack{\textbf{84.03}\\[-1pt]{\scriptsize$\mathbf{\pm 2.30}$}}
& \shortstack{\textbf{79.67}\\[-1pt]{\scriptsize$\mathbf{\pm 0.53}$}}
& \shortstack{\textbf{99.47}\\[-1pt]{\scriptsize$\mathbf{\pm 0.02}$}}
& \shortstack{\textbf{5.92}\\[-1pt]{\scriptsize$\mathbf{\pm 0.14}$}}
& \shortstack{\textbf{54.78}\\[-1pt]{\scriptsize$\mathbf{\pm 3.09}$}}
& \shortstack{\textbf{176.93}\\[-1pt]{\scriptsize$\mathbf{\pm 3.73}$}} \\
\midrule
\addlinespace[2pt]

Mem517~\citep{Mem517}
& 47.33 & 81.51 & 80.11 & 99.43 & 6.69 & 59.78 & 188.01 \\

MemFusion~\citep{MemFusion}
& 47.33 & 82.73 & 79.71 & 99.36 & 6.36 & 57.93 & 183.64 \\

MemMachine~\citep{MemMachine}
& 47.33 & 81.23 & 79.18 & 99.37 & 6.62 & 57.68 & 191.59 \\

Hybrid Memory Search~\citep{HybridSearchV2}
& 47.33 & 82.56 & 78.79 & 99.43 & 6.43 & 58.39 & 193.92 \\

\addlinespace[2.5pt]

EverOS~\citep{EverMemOS}
& 46.67 & 80.09 & 78.79 & 99.38 & 6.40 & 55.39 & 184.16 \\

ActiveMemoryIndex~\citep{ActiveMemoryIndex}
& 46.67 & 79.95 & 78.52 & 99.47 & 6.08 & 54.77 & 180.20 \\

\addlinespace[2.5pt]

Mi-Memory~\citep{MiMemory}
& 46.00 & 78.95 & 78.79 & 99.44 & 5.47 & 55.29 & 177.15 \\

\end{longtable}
\endgroup
% One recorded trajectory connects the native mechanisms and later agent use.
\newcommand{\memorycode}[1]{{\fontencoding{T1}\fontfamily{phv}\fontsize{9}{11}\selectfont #1}}
\section{Memory System Mechanisms}
\label{app:memory_representations}

This appendix compares how five memory systems represent an agent's solving history and retrieve relevant memories as context for Claude Code on later tasks.
Table~\ref{tab:memory_system_config} summarizes the implementations, internal models, and retrieval configurations used in our evaluation.

\begin{table}[!htbp]
\centering
\caption{Memory-system configurations used in our evaluation.}
\label{tab:memory_system_config}
\footnotesize
\setlength{\tabcolsep}{0pt}
\renewcommand{\arraystretch}{1.16}

\begin{tabular}{@{}
>{\raggedright\arraybackslash}p{0.13\linewidth}
@{\hspace{0.018\linewidth}}
>{\raggedright\arraybackslash}p{0.17\linewidth}
@{\hspace{0.018\linewidth}}
>{\raggedright\arraybackslash}p{0.14\linewidth}
@{\hspace{0.018\linewidth}}
>{\raggedright\arraybackslash}p{0.205\linewidth}
@{\hspace{0.018\linewidth}}
>{\raggedright\arraybackslash}p{0.265\linewidth}
@{}}
\toprule
\textbf{System} &
\textbf{Implementation} &
\textbf{Memory LLM} &
\textbf{Embedding Model} &
\textbf{Retrieval Configuration} \\
\midrule

A-Mem &
\texttt{ceffb860f0712}; MCP 0.2.6 (\texttt{dce0ac62cba5}) &
DeepSeek-V4-Flash &
all-MiniLM-L6-v2 (SentenceTransformers, Chroma) &
\texttt{k=5} for standard and agentic search; \texttt{k=10} for temporal search \\

\addlinespace[2pt]

Mem0 &
Plugin 0.2.9 (\texttt{366945965df4}); official API &
Provider-managed &
Provider-managed &
Official API defaults \\

\addlinespace[2pt]

MemOS &
MCP 1.1.2; official API &
Provider-managed &
Provider-managed &
\texttt{memory\_limit\_number=6}; retry with \texttt{relativity=0} when no result is returned \\

\addlinespace[2pt]

Supermemory &
Plugin 0.0.13 (\texttt{193290b5ccba}); official API &
Provider-managed &
Provider-managed &
\texttt{limit=5}; \texttt{searchMode=hybrid}; \texttt{threshold=0.6} \\

\addlinespace[2pt]

Claude-Mem &
v12.1.1 (\texttt{0b6997e3d03e}) &
DeepSeek-V4-Flash &
all-MiniLM-L6-v2 (ONNX, 384-d) &
\texttt{search limit=20}; 100 semantic candidates \\

\bottomrule
\end{tabular}

\vspace{2pt}
\parbox{\linewidth}{\footnotesize
\textit{Provider-managed} denotes components managed by the official API for which no fixed model identifier is exposed to the client.}
\end{table}

\noindent\textbf{Example task.}
The agent is asked to add a \memorycode{keep\_attrs} option to \memorycode{xarray.where}, allowing users to choose whether to preserve array metadata such as units and descriptions.
We use task \memorycode{pydata\_\_xarray-4687} from xarray, a data-processing library.
Its \memorycode{where} operation retains or replaces array values according to a condition.
At the time of this task, the \memorycode{where} methods on arrays and datasets (\memorycode{DataArray.where} and \memorycode{Dataset.where}) automatically preserved metadata.
However, the standalone function \memorycode{xarray.where} offered no option to control metadata preservation.

\begin{figure}[!htbp]
    \centering
    \includegraphics[width=\linewidth]{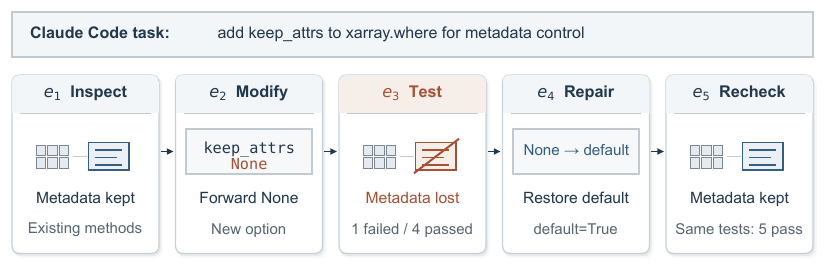}
    \caption{\textbf{Shared trajectory.} An initial change breaks default metadata preservation in the existing methods. The same five tests expose the regression and pass after the repair.}
    \label{fig:memory_trajectory_example}
\end{figure}

% Keep the trajectory introduction with the beginning of its event list.
\Needspace{5\baselineskip}
\noindent\textbf{Solving trajectory.}
Figure~\ref{fig:memory_trajectory_example} shows a segment of the agent's trajectory while solving this task, comprising five events:
\begin{enumerate}[label=$e_{\arabic*}$:,leftmargin=1.8em,labelsep=0.45em,itemsep=2pt,topsep=3pt,parsep=0pt,partopsep=0pt]
    \item \textbf{Inspect the existing behavior.} The agent inspects the existing \memorycode{where} methods and confirms that they automatically preserve metadata.
    \item \textbf{Add the option and modify the implementation.} While adding the new option, the agent also modifies the existing methods' internal implementation so that metadata preservation depends on the supplied \memorycode{keep\_attrs} value.
    \item \textbf{Detect a regression through testing.} The initial change breaks the existing default behavior. When \memorycode{keep\_attrs} is left unspecified, metadata is unexpectedly lost, causing a test to fail.
    \item \textbf{Correct the default handling.} The agent corrects how an unspecified \memorycode{keep\_attrs} value is handled, restoring metadata preservation by default.
    \item \textbf{Verify the repair.} The agent reruns the same set of \memorycode{where} tests, and all five tests pass.
\end{enumerate}

\noindent\textbf{Later retrieval.}
When Claude Code encounters a related metadata-preservation issue in a later task, this history can provide the earlier failure's cause, repair, and validation.
To illustrate memory storage and retrieval, we use a shared example query, $q$: ``How should \memorycode{where} preserve attributes?''
The following five system figures show possible memory representations of this history.
They also illustrate how each system retrieves relevant context for Claude Code in response to $q$.
The $e_i$ labels identify source events.
A memory may integrate information from multiple events.

\FloatBarrier

\subsection{A-Mem: Evolving Note Graph}
\label{app:memory_amem}

\noindent\textbf{Memory construction.}
A-Mem~\citep{AMem} organizes reusable knowledge from Claude Code as linked notes.
In this integration, a session-start hook injects instructions to search memory before each task and save new findings through MCP during code exploration, debugging, or repair.
The \memorycode{add\_memory\_note} description covers architectural relations, code locations, errors and solutions, recurring implementation patterns, and configuration requirements.
It excludes trivial or obvious information, findings already covered by memory, and observations without reusable insights.

\begin{figure}[!htbp]
    \centering
    \includegraphics[width=\linewidth]{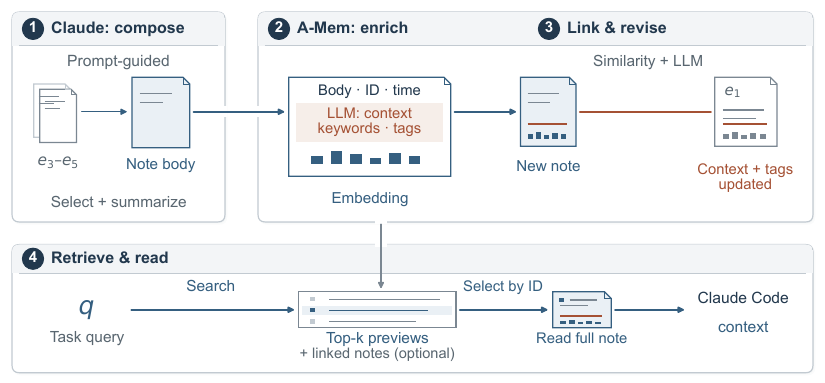}
    \caption{\textbf{A-Mem with Claude Code.} Claude composes a note from $e_3$--$e_5$, which A-Mem enriches and links. Retrieval combines semantic previews, selective reading, and optional link expansion.}
    \label{fig:memory_workflow_amem}
\end{figure}

In Step~1 of Figure~\ref{fig:memory_workflow_amem}, Claude selects new findings from its current task context and summarizes them as a self-contained note body.
It submits this text as the tool's \memorycode{content} argument.
Claude authors the body from its exploration, edits, and test results.
A-Mem receives this text rather than extracting it from the full trajectory.
Here, a note can combine metadata loss ($e_3$), repair ($e_4$), and a passing recheck ($e_5$) to record the failure conditions, solution, and validation.

In Step~2, A-Mem creates a note containing this body, a unique identifier, and a timestamp.
Its LLM extracts keywords for key concepts, assigns semantic tags, and generates a one-sentence \memorycode{context} description of the topic, main points, and purpose.
These fields form the note's semantic attributes.
A text encoder embeds the concatenated body and attributes for indexing in the vector database.

In Step~3, A-Mem encodes the new note's body with the same text encoder.
It compares this query vector with stored embeddings of note bodies and semantic attributes to retrieve the nearest candidates.
An LLM compares the candidates' bodies and attributes with the new note to select links.
A-Mem records the selected identifiers as links.
It can also update existing \memorycode{context} descriptions and tags while retaining the original bodies.
Here, the repair note can link to the note from $e_1$.
The earlier note's description can then explain how to handle an unspecified \memorycode{keep\_attrs} value.

\noindent\textbf{Memory retrieval.}
The hook prompts Claude to retrieve memory, but does not query the database itself.
Claude formulates query $q$ from the current task and submits it through \memorycode{search\_memories}.
In Step~4, A-Mem encodes $q$ with the same encoder and returns the top-$k$ nearest notes as lightweight previews: identifiers, \memorycode{context}, keywords, tags, and distance scores.
Claude assesses their relevance and calls \memorycode{read\_memory\_note} to read selected bodies and links by identifier.
For broader context, \memorycode{search\_memories\_agentic} extends the retrieval pool with linked neighbors before returning results for further reading.
This combination of semantic search and selective reading lets Claude locate prior experience, then inspect its failure conditions, repair, and validation.

\FloatBarrier

\subsection{Mem0: Persistent Repository Memory}
\label{app:memory_mem0}

\begin{figure}[!htbp]
    \centering
    \includegraphics[width=\linewidth]{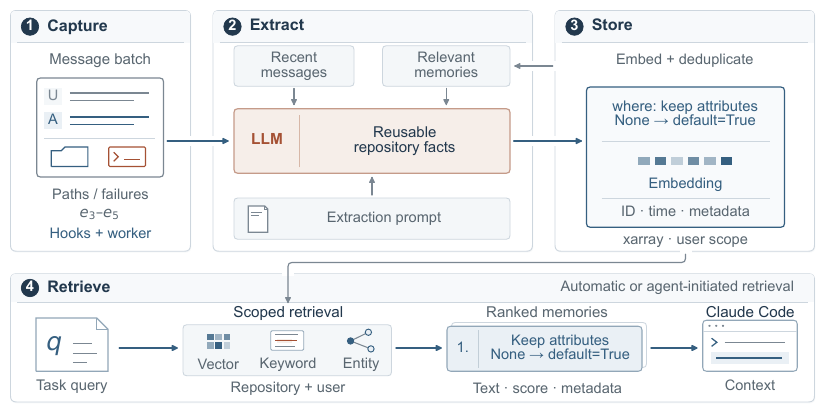}
    \caption{\textbf{Mem0 with Claude Code.} Selected interaction evidence is distilled into persistent memories for scoped retrieval.}
    \label{fig:memory_workflow_mem0}
\end{figure}

\noindent\textbf{Memory construction.}
Mem0~\citep{Mem0} distills an agent's interaction history into persistent memory for retrieval across sessions.
In Step~1 of Figure~\ref{fig:memory_workflow_mem0}, Claude Code lifecycle hooks apply predefined rules to collect user prompts, agent responses, and selected execution evidence, including modified file paths and bounded diagnostics for failed commands.
The plugin accumulates this material locally, and a background worker submits it to Mem0 in message batches.
The evidence payload excludes full file contents and general tool output.

Step~2 illustrates the current open-source text-memory extraction pipeline.
The submitted \memorycode{user} and \memorycode{assistant} messages, together with the selected evidence, provide the primary input.
Recent messages within the session scope supply context and help resolve references.
Before extraction, Mem0 uses the current messages to retrieve semantically related records within the same scope, helping the LLM distinguish new information from existing knowledge.
A dedicated extraction prompt then guides an LLM to retain established repository behavior, relevant implementation locations, important constraints, and reusable solutions.
It excludes transient exploration, unaccepted proposals, and standalone test results.
The LLM produces concise memory entries, or none when no useful information has been established.
For $e_3$--$e_5$, an entry can state that \memorycode{where} should preserve attributes when \memorycode{keep\_attrs} is unspecified, resolving the default through \memorycode{\_get\_keep\_attrs(default=True)}.

Step~3 persists the extracted memories.
Mem0 computes their embeddings, filters duplicate entries, and stores each record with its text, identifier, timestamps, scope, and metadata.
The \memorycode{where} rule thus becomes a record associated with the xarray repository.
Its text captures reusable repository knowledge, its embedding supports similarity search, and its scope limits retrieval and reuse to relevant tasks.

\noindent\textbf{Memory retrieval.}
Step~4 returns persistent memories to Claude Code during subsequent tasks.
The first substantive user prompt in a new session initiates a search within the applicable repository and user scope.
During later execution, the agent can also call \memorycode{search\_memories} with a natural-language query about repository behavior, an error, or an implementation detail.
Given $q$, the platform combines similarity between query and memory embeddings with keyword and entity matching to rank eligible records.
Temporal metadata can further restrict results through filters.
The search returns the highest-ranked memories with relevance scores and metadata, and the plugin provides selected memory text to the agent as compact context.

\FloatBarrier

\subsection{MemOS: Text Memories in MemCube}
\label{app:memory_memos}

\begin{figure}[!htbp]
    \centering
    \includegraphics[width=\linewidth]{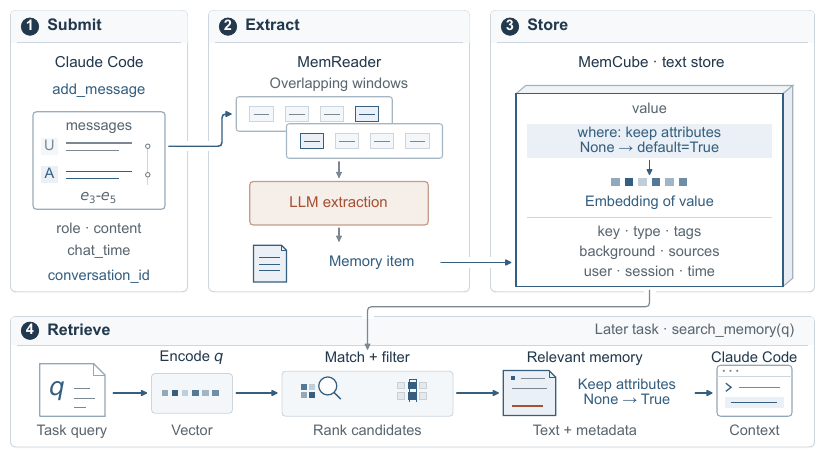}
    \caption{\textbf{MemOS with Claude Code.} MemReader extracts textual memories into MemCube for later semantic retrieval.}
    \label{fig:memory_workflow_memos}
\end{figure}

\noindent\textbf{Memory construction.}
MemCube is the unified memory abstraction proposed by MemOS and implemented in its open-source framework as a modular container for textual, activation, and parameter memories~\citep{MemOS}.
It supports the organization and persistence of these memory types.
Figure~\ref{fig:memory_workflow_memos} follows the textual path, which converts Claude Code's past interactions into structured records for storage and later reuse.

In Step~1, Claude Code submits past interactions through the MCP tool \memorycode{add\_message}.
The interface associates an ordered \memorycode{messages} array with a \memorycode{conversation\_id} before forwarding it to the service.
Each message carries \memorycode{role} and \memorycode{content}, with \memorycode{chat\_time} supplied by the caller or filled in by the interface.
User and assistant exchanges associated with $e_3$--$e_5$ thus provide the input to memory extraction.

Step~2 follows MemReader's LLM extraction path in the open-source framework.
Consecutive messages are grouped into context windows bounded by a token limit.
Adjacent windows may overlap to preserve context across turns.
For each window, an LLM extracts reusable facts, decisions, and experience as structured memory items, including a memory statement (\memorycode{value}), a short identifying \memorycode{key}, a memory type, and tags.
Window background and supporting messages are retained.
For $e_3$--$e_5$, the extracted statement can specify that the existing \memorycode{where} methods should resolve \memorycode{keep\_attrs=None} using \memorycode{\_get\_keep\_attrs(default=True)} to preserve attributes by default.

Step~3 shows how these items are represented and persisted in MemCube's text store.
When preparing each record, the embedding model encodes only \memorycode{value} for semantic retrieval.
The \memorycode{key}, memory type, tags, background, and source messages are retained as metadata, together with user, session, and time information.
The record therefore preserves both the text used for semantic matching and the contextual attributes stored alongside it.

\noindent\textbf{Memory retrieval.}
In subsequent tasks, Claude Code formulates $q$ from the current problem and calls \memorycode{search\_memory}.
Step~4 illustrates query encoding, semantic matching against embeddings derived from the stored \memorycode{value} text, and candidate filtering and ranking.
Metadata filters constrain the eligible records, while semantic relevance guides selection.
When attribute handling in \memorycode{where} becomes relevant again, the experience extracted from $e_3$--$e_5$ can be returned to Claude Code as context, providing the earlier behavioral constraint and implementation rule.

\FloatBarrier

\subsection{Supermemory: Documents, Memory Graphs, and Profiles}
\label{app:memory_supermemory}

\begin{figure}[!htbp]
    \centering
    \includegraphics[width=\linewidth]{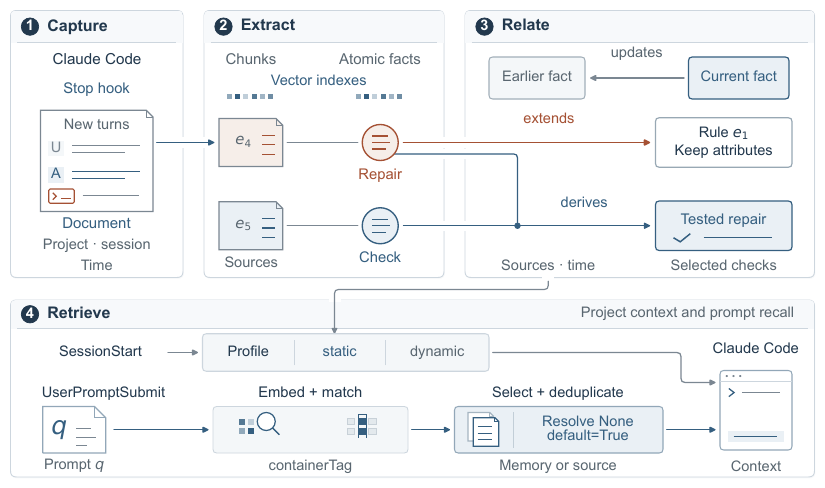}
    \caption{\textbf{Supermemory with Claude Code.} Documents yield linked atomic memories for project profiles and subsequent retrieval.}
    \label{fig:memory_workflow_supermemory}
\end{figure}

\noindent\textbf{Memory construction.}
Supermemory stores Claude Code interactions as source documents and distills them into retrievable atomic memories~\citep{SuperMemory}.
In Step~1 of Figure~\ref{fig:memory_workflow_supermemory}, the \memorycode{Stop} hook reads the session transcript after Claude completes a response, selecting entries added since the last successful submission.
It preserves the order of user messages, assistant responses, and configured tool calls and results, then submits this increment as document content with project, session, and time information.

In Step~2, Supermemory partitions documents into coherent chunks and indexes their embeddings.
Extraction instructions supplied by the plugin prioritize reusable decisions, conventions, workflows, and lessons.
Atomic memories isolate individual facts while resolving the objects, operations, and parameters needed to interpret them.
Each memory retains links to its supporting chunks and an embedding for retrieval.
For example, $e_4$ can yield the rule for resolving \memorycode{keep\_attrs=None} in the existing \memorycode{where} methods, while $e_5$ records the corresponding focused-test result.
Memory generation may continue after chunks become searchable.

In Step~3, these facts join the memory graph with provenance and temporal metadata.
An \emph{updates} relation makes a corrected fact current while retaining the earlier version as history.
An \emph{extends} relation adds compatible details while both memories remain valid.
A \emph{derives} relation links an inferred conclusion to its premises.
Thus, $e_4$ can extend $e_1$'s attribute-preservation rule, and $e_4$ together with $e_5$ can support the conclusion that the repair passed the selected checks.
The service also maintains a profile for the project, whose \memorycode{static} component summarizes stable facts and whose \memorycode{dynamic} component captures recent activity and evolving context.

\noindent\textbf{Memory retrieval.}
At session start, the \memorycode{SessionStart} hook retrieves the project's \memorycode{static} and \memorycode{dynamic} profile components and supplies them to Claude Code.
For subsequent substantive prompts, \memorycode{UserPromptSubmit} requests matching content using the current prompt as $q$ and the project's \memorycode{containerTag}.
In Step~4, the query is embedded for semantic matching.
The service can search memory embeddings, document-chunk embeddings, or both, with metadata constraints and optional relation expansion or reranking.
The hook filters low-scoring matches, removes text already recalled during the session, and injects concise context into Claude Code.
For $q$, matching memory or source text can expose $e_4$'s default-resolution rule and $e_5$'s focused verification.

\FloatBarrier

\subsection{Claude-Mem: Observations and Progress Summaries}
\label{app:memory_claudemem}

\begin{figure}[!htbp]
    \centering
    \includegraphics[width=\linewidth]{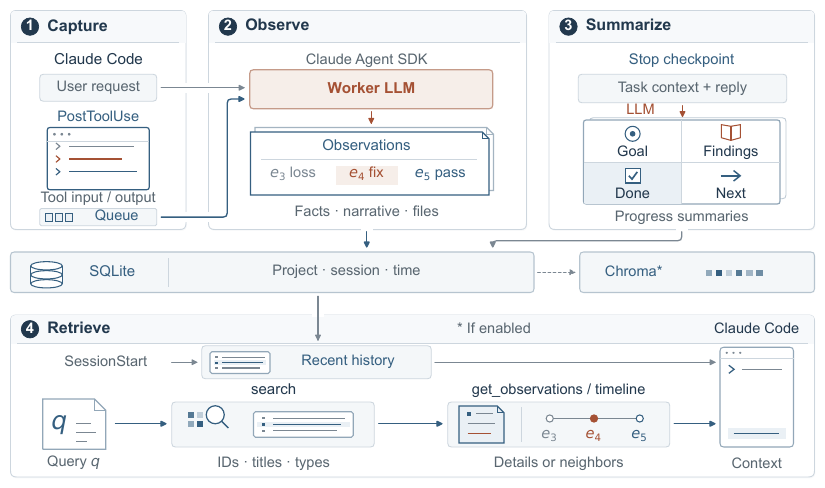}
    \caption{\textbf{Claude-Mem with Claude Code.} Observations and progress summaries support startup context and selective retrieval.}
    \label{fig:memory_workflow_claudemem}
\end{figure}

\noindent\textbf{Memory construction.}
Claude-Mem captures coding activity through Claude Code's lifecycle hooks and compresses it into observations and progress summaries in a background worker~\citep{ClaudeMem}.
In Step~1 of Figure~\ref{fig:memory_workflow_claudemem}, the user request establishes task context.
After each tool call, \memorycode{PostToolUse} forwards the tool name, arguments, result, and working directory to a background queue.
Code reads, edits, and test runs thus supply raw activity records for subsequent extraction.

In Step~2, the worker's Claude Agent SDK backend invokes an LLM with the user request as context and the current tool's name, input, and result as evidence.
The extraction prompt permits skipping uninformative activity or producing multiple observations.
Each observation contains a type, title, \memorycode{facts}, an explanatory \memorycode{narrative}, concepts, and files read or modified.
For example, observations can record $e_3$'s metadata loss, $e_4$'s default-resolution repair, and $e_5$'s focused-test result.

Step~3 maintains a complementary session-level record.
The \memorycode{Stop} hook requests a progress summary using the worker's accumulated task context and Claude's latest response.
The LLM summarizes the request, investigation, lessons, completed work, next steps, and notes, allowing successive checkpoints within one session.
Observations and summaries are stored in SQLite with project, session, and time metadata.
When Chroma indexing is enabled, an observation's \memorycode{narrative} and individual \memorycode{facts} receive separate vector representations linked to its SQLite ID.

\noindent\textbf{Memory retrieval.}
At session start, \memorycode{SessionStart} uses recent observations and summaries from the project's SQLite records to construct a compact chronological overview for Claude Code.
Selection follows project, time, and configured context filters without requiring semantic search.
The overview exposes earlier work and progress before the agent requests further details.

In Step~4, Claude can query history through MCP tools.
For observation retrieval, \memorycode{search} returns compact candidates with IDs, titles, types, and timestamps.
With Chroma enabled, query similarity matches indexed narratives and facts, while metadata filters constrain project, type, concepts, files, or date.
Claude can then use \memorycode{get\_observations} to read complete facts, narratives, and file information for selected IDs, or \memorycode{timeline} to inspect neighboring events around an anchor ID or query.
For $q$, this can locate $e_4$'s repair and place it alongside the preceding failure and $e_5$'s verification.

\FloatBarrier

\section{Metric Definitions}
\label{app:metrics}

This appendix gives the full definitions of the evaluation metrics used in the
paper. Let $N$ denote the number of base tasks in an evaluation condition.

\subsection{Main Evaluation Metrics}
\label{app:main_metrics}
\textbf{Resolved.}
Resolved measures the fraction of base tasks solved according to the official
task verifier. For a single run,
\begin{equation}
\mathrm{Resolved}
=
\frac{1}{N}
\sum_{i=1}^{N}
\mathbb{I}[V_i = 1],
\end{equation}
where $V_i=1$ if the agent's final patch for task $i$ passes the verifier and
$V_i=0$ otherwise. Resolved is therefore an all-or-nothing measure of
task-solving success.

\textbf{Rescue and Harm.}
The change in Resolved relative to NoMem captures only the net effect of memory.
Rescue and Harm separate this change into positive and negative effects:
Rescue captures cases where NoMem fails but the condition with memory succeeds,
whereas Harm captures the reverse.

Each condition is evaluated with three independent runs. For task $i$, let
$k_i \in \{0,1,2,3\}$ denote the number of successful NoMem runs and
$m_i \in \{0,1,2,3\}$ the number of successful runs under the memory condition.
Because runs across conditions have no natural one-to-one correspondence, we
consider all $3\times3$ cross-condition run pairs for each task. Among these
nine pairs, $(3-k_i)m_i$ have a failed NoMem run and a successful memory run,
while $k_i(3-m_i)$ exhibit the reverse. Rescue and Harm are therefore defined as
\begin{equation}
\mathrm{Rescue}
=
\frac{1}{9N}
\sum_{i=1}^{N}(3-k_i)m_i,
\qquad
\mathrm{Harm}
=
\frac{1}{9N}
\sum_{i=1}^{N}k_i(3-m_i).
\end{equation}

Their difference exactly recovers the change in Resolved relative to NoMem:
\begin{equation}
\mathrm{Rescue} - \mathrm{Harm}
=
\frac{1}{3N}
\sum_{i=1}^{N}(m_i-k_i)
=
\mathrm{Resolved}_{\mathrm{system}}
-
\mathrm{Resolved}_{\mathrm{NoMem}}.
\end{equation}

For Rescue and Harm, we compute $95\%$ confidence intervals by bootstrapping
over tasks. Each bootstrap replicate samples $N$ tasks with replacement while
retaining all three outcomes under each condition, and recomputes Rescue and
Harm using the same $3\times3$ cross-condition pairing. We repeat this process
$10{,}000$ times and report the $2.5$th and $97.5$th percentiles of the resulting
bootstrap distributions.

\textbf{Input Tokens and Output Tokens.}
Input Tokens and Output Tokens measure the token cost of base-task solving.
They exclude the tokens consumed while memory systems build memory from historical tasks.
For task $i$, let $T_i^{\mathrm{in}}$ and $T_i^{\mathrm{out}}$ denote the
total numbers of input and output tokens consumed during solving. We compute
the per-task means
\begin{equation}
\mathrm{InputTokens}
=
\frac{1}{N}
\sum_{i=1}^{N}T_i^{\mathrm{in}},
\qquad
\mathrm{OutputTokens}
=
\frac{1}{N}
\sum_{i=1}^{N}T_i^{\mathrm{out}}.
\end{equation}

\textbf{Agent Steps.}
Agent Steps measures the number of model-response turns produced by the agent
during base-task solving. Each model response is counted as one step and may
contain reasoning, textual output, or tool calls; multiple tool calls within
the same response are counted as a single step. Let $S_i$ denote the total
number of steps for task $i$. We compute
\begin{equation}
\mathrm{AgentSteps}
=
\frac{1}{N}
\sum_{i=1}^{N}S_i.
\end{equation}

\subsection{Partial Repair Metrics}
\label{app:repair_metrics}

\textbf{F2P and P2P.}
Resolved credits a task only when all required tests pass and therefore does
not capture partial repair progress. For task $i$, let $F_i$ denote its
fail-to-pass tests, which fail on the original repository state and should pass
after a correct repair, and let $P_i$ denote its pass-to-pass tests, which pass
on the original state and should remain passing. Both sets come from the
official task specification.
Let $f_i \le |F_i|$ denote the number of fail-to-pass tests that pass after
the agent's final patch is applied, and let $p_i \le |P_i|$ denote the number
of pass-to-pass tests that remain passing. We micro-average over tests:
\begin{equation}
\mathrm{F2P}
=
\frac{\sum_{i=1}^{N} f_i}
     {\sum_{i=1}^{N} |F_i|},
\qquad
\mathrm{P2P}
=
\frac{\sum_{i=1}^{N} p_i}
     {\sum_{i=1}^{N} |P_i|}.
\end{equation}
F2P measures how much of the requested behavior is implemented, whereas P2P
measures how well previously passing behavior is preserved.

\subsection{Gold-Patch Alignment Metrics}
\label{app:patch_metrics}

Resolved, F2P, and P2P measure the agent's functional performance through test
outcomes. As a complement, we further compare the agent's final patch with the
reference solution patch, measuring their agreement at the file, edit-location,
and edit-content levels.

\textbf{Patch Sets.}
For task $i$, let $G_i$ denote the set of files modified by the reference
solution patch, excluding the test patch, and let $C_i^{r}$ denote the set of
files modified by the agent's final patch in run $r$. For all three metrics, we
first average over tasks within each run and then report the mean and sample
standard deviation across three runs.

\textbf{File Recall.}
File Recall measures how many files modified by the reference solution patch are
covered by the agent. It is defined as the fraction of reference files that also
appear in the agent's modified-file set:
\begin{equation}
\mathrm{FileRecall}_i^{r}
=
\frac{|C_i^{r} \cap G_i|}{|G_i|}.
\end{equation}

% Keep the short operation definitions with their introductory paragraph.
\Needspace{13\baselineskip}
\textbf{Line Recall.}
Line Recall further measures whether the agent edits the locations modified by
the reference solution patch. We treat each added or deleted line in a unified
diff as one edit operation and ignore context lines; a line replacement
corresponds to one deletion and one addition. A deletion is represented as
\[
(\mathrm{file}, \texttt{delete}, \ell, 0),
\]
where $\ell$ is the line number in the pre-edit file. An addition is represented
as
\[
(\mathrm{file}, \texttt{add}, a, o),
\]
where $a$ is its insertion anchor in the pre-edit file and $o$ is its offset
among additions sharing that anchor.

Let $L_i$ and $A_i^{r}$ denote the operation sets of the reference solution
patch and the agent's final patch, respectively. Line Recall is defined as the
fraction of reference operations whose locations are matched by the agent:
\begin{equation}
\mathrm{LineRecall}_i^{r}
=
\frac{|A_i^{r} \cap L_i|}{|L_i|}.
\end{equation}
An operation is matched only when the file, edit direction, location, and offset
all agree. Line Recall does not compare the changed text.

% Keep this short definition, its tuple, and its equation together.
\Needspace{15\baselineskip}
\textbf{Edit Recall.}
Edit Recall further requires the changed text to match at the same location. We
extend each operation to
\[
(\mathrm{file}, \texttt{add}/\texttt{delete},
\text{anchor or line}, \text{offset}, \mathrm{rstrip}(t)),
\]
where $t$ is the changed-line text and $\mathrm{rstrip}$ removes trailing
whitespace only, preserving leading indentation, internal whitespace, and case,
with no other normalization.

Let $E_i$ and $D_i^{r}$ denote the resulting operation sets for the reference
solution patch and the agent's final patch, respectively. Edit Recall is defined
as the fraction of reference operations whose location and text are both
matched exactly:
\begin{equation}
\mathrm{EditRecall}_i^{r}
=
\frac{|D_i^{r} \cap E_i|}{|E_i|}.
\end{equation}
An operation is matched only when the entire tuple agrees exactly.

\end{document}